\documentclass[final,1p,times,twocolumn]{elsarticle}
\usepackage{graphicx,subcaption}
\usepackage{lineno,hyperref}
\modulolinenumbers[5]
\usepackage[section]{placeins}
\journal{Journal of Mechanical Systems and Signal Processing}
\usepackage{booktabs}
\usepackage{graphicx} 
\usepackage{wrapfig}%package to manage images
\usepackage{ragged2e}
 \usepackage{cleveref}
\usepackage[utf8]{inputenc}
\usepackage[english]{babel}
\usepackage{fancyhdr}
\usepackage{booktabs}
 \usepackage{nopageno}
\usepackage{caption}
\usepackage[font=normalsize]{caption} 
\addto\captionsenglish{}
\usepackage{nopageno}
\usepackage{wrapfig}
\usepackage{caption}
\usepackage{threeparttable}
\usepackage[ruled,vlined]{algorithm2e}
\usepackage[font=normalsize]{caption} 
\addto\captionsenglish{}
\SetKwInput{KwInput}{Input}                % Set the Input
\SetKwInput{KwOutput}{Output} 
\begin{document}
\begin{frontmatter}

\title{System-reliability based multi-ensemble of GAN and one-class joint Gaussian distributions for unsupervised real-time structural health monitoring}

%% Group authors per affiliation:
\author[1]{Mohammad Hesam Soleimani-Babakamali}
\author[1,2]{Reza Sepasdar}
\author[3]{Kourosh Nasrollahzadeh}
\author[1]{Rodrigo Sarlo\corref{mycorrespondingauthor}}
\address[1]{Department of Civil and Environmental Engineering, Virginia Tech University, Blacksburg, VA, USA}
\address[2]{Department of Computer Science, Virginia Tech University, Blacksburg, VA, USA}
\address[3]{Department of Civil Engineering, K. N. Toosi University of Technology, Tehran, Iran}
%\fntext[myfootnote]{Since 1880.}

%% or include affiliations in footnotes:
%\author[mymainaddress,mysecondaryaddress]{Elsevier Inc}
%\ead[url]{www.elsevier.com}

%\author[mysecondaryaddress]{Global Customer Service\corref{mycorrespondingauthor}}
\cortext[mycorrespondingauthor]{Corresponding author, Sarlo@vt.edu}
%\ead{Soleimanihesam92@vt.edu}

%\address[mymainaddress]{1600 John F Kennedy Boulevard, Philadelphia}
%\address[mysecondaryaddress]{360 Park Avenue South, New York}
%\FloatBarrier
\begin{abstract}

Unsupervised health monitoring has gained much attention in the last decade as the most practical real-time structural health monitoring (SHM) approach. Among the proposed unsupervised techniques in the literature, there are still obstacles to robust and real-time health monitoring. These barriers include loss of information from dimensionality reduction in feature extraction steps, case-dependency of those steps, lack of a dynamic clustering, and detection results' sensitivity to user-defined parameters. This study introduces an unsupervised real-time SHM method with a mixture of low- and high-dimensional features without a case-dependent extraction scheme. Both features are used to train multi-ensembles of Generative Adversarial Networks (GAN) and one-class joint Gaussian distribution models (1-CG). A novelty detection system of limit-state functions based on GAN and 1-CG models' detection scores is constructed. The Resistance of those limit-state functions (detection thresholds) is tuned to user-defined parameters with the GAN-generated data objects by employing the Monte Carlo histogram sampling through a reliability-based analysis. The tuning makes the method robust to user-defined parameters, which is crucial as there is no rule for selecting those parameters in a real-time SHM. The proposed novelty detection framework is applied to two standard SHM datasets to illustrate its generalizability: Yellow Frame (twenty damage classes) and Z24 Bridge (fifteen damage classes). All different damage categories are identified with low sensitivity to the initial choice of user-defined parameters with both introduced dynamic and static baseline approaches with few or no false alarms.

\end{abstract}

\begin{keyword}
\texttt Unsupervised real-time SHM; Generative adversarial networks; Gaussian mixture models; Anomaly detection, System-reliability; Monte Carlo histogram sampling
\end{keyword}

\end{frontmatter}

% \linenumbers
%\FloatBarrier
\section{Introduction}
\label{Int}

Cities are a mixture of different structures, such as buildings, bridges, and other physical infrastructure. Incorporating Structural Health Monitoring (SHM) into such a vast network is vital to save resources, reduce repair costs with timely maintenance, and save human lives. This task can be accomplished via inexpensive and highly-accurate sensors and robust novelty detection frameworks. A novelty in a structure is a change in the acquired sensors' data resulting from damage in the system or a shift in the structure's response due to environmental factors. Generally, the SHM task is accomplished via either model-based or data-driven approaches. In model-based approaches, typically, a finite element model is fitted or updated, based on experimental sensor data \cite{hua2009structural,umar2018response}. By regularly updating a model, possible damages and their locations can be discovered. However, these methods rely on complex optimization algorithms and dense sensor deployments for capturing many mode shapes, including local modes, \cite{entezami2018unsupervised}. Furthermore, the need for the presence of an expert and the vulnerability of detection results to noisy data \cite{hua2009structural,umar2018response} make this type of approach unsuitable for real-time SHM with noisy data. Moreover, missing a novelty is probable if the model's physics cannot capture the effect of a novelty \cite{avci2020review}. Some improvements are achieved by assuming probabilistic model parameters \cite{entezami2018unsupervised,zhao2020bayesian}; however, the proposed methods are mostly validated versus numerical simulations and laboratory testings and need further field experiments to prove their validity \cite{zhao2020bayesian}.

Data-driven approaches, including supervised, unsupervised, and semi-supervised strategies, are alternatives to the model-based techniques. A typical example of a supervised method is applying subspace identification while using the angles between subspaces as damage indicators \cite{yan2006null}. Another example is the use of 1-D or 2-D convolutional neural network layers for damage classification \cite{abdeljaber20181,yu2019novel}, shown to work well even in noisy situations. However, supervised methods are not well suited to the real-time SHM of structures. Gathering enough data objects of different structural conditions from facilities in operation is costly and impractical in most cases. Some efforts have been made to alleviate this issue, such as having only two classes of an undamaged and fully damaged system for performing the damage detection \cite{abdeljaber20181}, or generating data objects from low-sampled damage classes via 1-D Generative adversarial networks (GAN) to improve the robustness of classifiers \cite{soleiman2020}. Unfortunately, obtaining fully-damaged states is a big challenge in itself. The GAN-generated data objects can also help if there are data objects from a low-sampled class and cannot be applied for unrecorded damage classes.
 
Unlike the model-based and supervised techniques, unsupervised techniques are more naturally suited to real-time SHM because no prior information or assumptions of abnormal structural states is required. However, current approaches still have several challenges, particularly when applied across multiple structures. For conciseness, the current challenges in the state of the art studies in the unsupervised SHM field are investigated here. These challenges can be broadly classified into four categories: loss of information from dimensionality reduction (\textit{loss of information}), case-dependency of feature extraction methods (\textit{case-dependency}), lack of a dynamic clustering (\textit{non-dynamic clusters}), and the novelty detection's sensitivity to user-defined parameters (\textit{sensitivity to initial parameters}), each of which is described below.
 
The first challenge is the \textit{loss of information}. Most unsupervised learning studies involve huge dimensionality reduction to avoid the curse of dimensionality \cite{tan2016introduction} specifically for clustering. Although this makes clustering practical, the baseline condition's reduced representation may overlook dimensions important for describing unknown or unexpected situations. Some specific layers, such as one or two-dimensional convolutional layers, RNNs, and deep fully-connected layers, have been shown to classify (i.e., supervised) novelties in high-dimensional data. Such networks cannot be used in an unsupervised setting, as there is no classification. Still, some related approaches, such as GANs \cite{goodfellow2014generative}, can be applied to take advantage of those layers in an unsupervised manner. 2-D GAN was applied in unsupervised anomalous sensor output validation \cite{mao2020toward}. In that study, sensor outputs (i.e., time-series data) are transformed into Gramian Angular Field images to train a GAN model. The GAN's Generator is then embedded as the decoder of a deep autoencoder network, with the autoencoder's reconstruction loss as the novelty detection criterion. In SHM, on the other hand, 1-D data are of more interest (i.e., frequency or time-domain data series). The 1-D GAN idea for constructing lost sensor data \cite{lei2020lost}, and the classification improvement of low-sampled damage scenarios \cite{soleiman2020} are examples of 1-D GAN implementations in SHM.
 
The second challenge, \textit{case-dependency}, refers to the limitations in applying the feature extraction steps (i.e., high or low-dimensional) of current approaches to a new structure or across multiple facilities. For example, it is impractical to train deep auto-encoders \cite{pathirage2018structural,wang2020unsupervised,ma2020structural,wang2020unsupervised}, deep Boltzmann machines \cite{rafiei2018novel}, or Kernel-based PCA approaches \cite{langone2017automated} on all the structures in a network. The application of those techniques to a new structure from scratch is time-consuming. Defining a simple extraction technique can aid its implementation on any new structure. Still, such a simple extraction technique may need special tools (e.g., neural network architecture) to handle the resulting features.
 
The third challenge is \textit{non-dynamic clusters}. Having one-class novelty detection or support vector machines \cite{sarmadi2020novel, wang2020unsupervised}, or a user-defined static number of clusters before the detection starts \cite{bull2019probabilistic,bull2020towards} was performed in unsupervised SHM. However, one-class models cannot record and save different damage scenarios, and the correct number of classes to be unknown at the start of the detection phase. A dynamic-class approach can identify such trends, and if the same novelty returns in the future, it recognizes it without flagging a new novelty. Bouzenad \textit{et al.} \cite{bouzenad2019semi} defined a threshold for the distance of a new test data object with available clusters to increase clusters (i.e., dynamic-class) in case of the threshold being exceeded. However, the lack of a probabilistic framework to tune that threshold can reduce its effectiveness in practice, as different values of it can generate more or fewer clusters than the ground truth clusters \cite{bouzenad2019semi}.
 
The fourth challenge is the \textit{sensitivity to initial parameters}. User-defined (detection) parameters such as the number of time-series data object to undergo the detection process in each detection-iteration (i.e., detection window length) \cite{de2019automated}, or thresholds for having a dynamic number of clusters \cite{bouzenad2019semi} can result in a variety of clusters, or alarms differently from the ground truth ones. The detection window length is negatively correlated to the novelty detection power and positively correlated with having less novelty detection while the system is intact (false alarms) \cite{santos2015baseline,de2019automated}. In studies dealing with the detection window length, different lengths were studied 
\cite{santos2015baseline,de2019automated,entezami2018unsupervised}, and the one with the best detection result was introduced as the best candidate. Nevertheless, tuning the detection-window length is impractical in a real-time detection process with unseen data, as it must be defined before the initiation of the analysis. 
 
This study defines a pair of simple low- and high-dimensional arithmetic-based frequency-domain features applicable to any structure or across multiple structures to address \textit{loss of information} and \textit{case-dependency} challenges. The GAN model is decided for dealing with the high-dimensional feature, while the low-dimensional feature is fed into a one-class joint Gaussian distribution model (1-CG). Utilizing GAN, specifically its Discriminator, as a classifier in an unsupervised approach is a novel strategy in unsupervised real-time SHM literature, making it possible to have multi-GAN's Discriminators to mimic a dynamic-class novelty detection framework. A novelty detection system with several limit-state functions are defined based on the classification scores of the GAN's Discriminator and 1-CG model for the incoming data. With Generator-generated F\_I features initiated from a probability density function (a latent space), reliability-based approaches can be used to dynamically tunes the limit-state functions' Resistance (detection threshold) to user-defined parameters. The employment of reliability-based techniques, which is not tried before in unsupervised real-time SHM literature, is the main novelty of this study. By having the model parameters as random variables in model-based approaches, the reliability-based techniques are recently applied for model-based damage detection to derive a formula that incorporates the probability of damage detection and false alarms to define the minimum detectable damages \cite{MENDLER2021107561}. In the current study, the GAN's latent dimension, and the generated data objects from it, is the reason that enables a reliability analysis in data-driven SHM. The detection thresholds' tuning addresses the \textit{sensitivity to initial parameters}. The proposed novelty detection framework is tested on two datasets, Yellow Frame with twenty-one classes of normal and damaged states and Z-24 bridge on its last month before demolition, with multiple damage scenarios.
 
In what follows, both datasets are explained, and then different parts of the methodology are introduced. Furthermore, the GAN's network architecture is determined based on the proposed feature extraction step. From the trained GAN and 1-CG models, two scores are introduced for unsupervised real-time novelty detection. The novelty detection system is presented in the next step, and its detection thresholds are tuned through a reliability-based analysis method. Finally, in both datasets, one-class (static) and dynamic-class approaches are applied, with the results being discussed.

\section{Case studies}

For real-time SHM, and specifically for evaluating how the proposed framework is working, experimental datasets include multiple damage scenarios with defined damage initiation points are essential. In this section, two such datasets are introduced: Yellow Frame and Z24 Bridge. A summary of both datasets' structures, a brief description of their acquisition systems, and the datasets summaries are provided.

\subsection{Yellow Frame}

\begin{figure}[h!]
    \centering
    \begin{subfigure}[b]{0.48\linewidth} 
        \centering
     \includegraphics[width=0.75\textwidth]{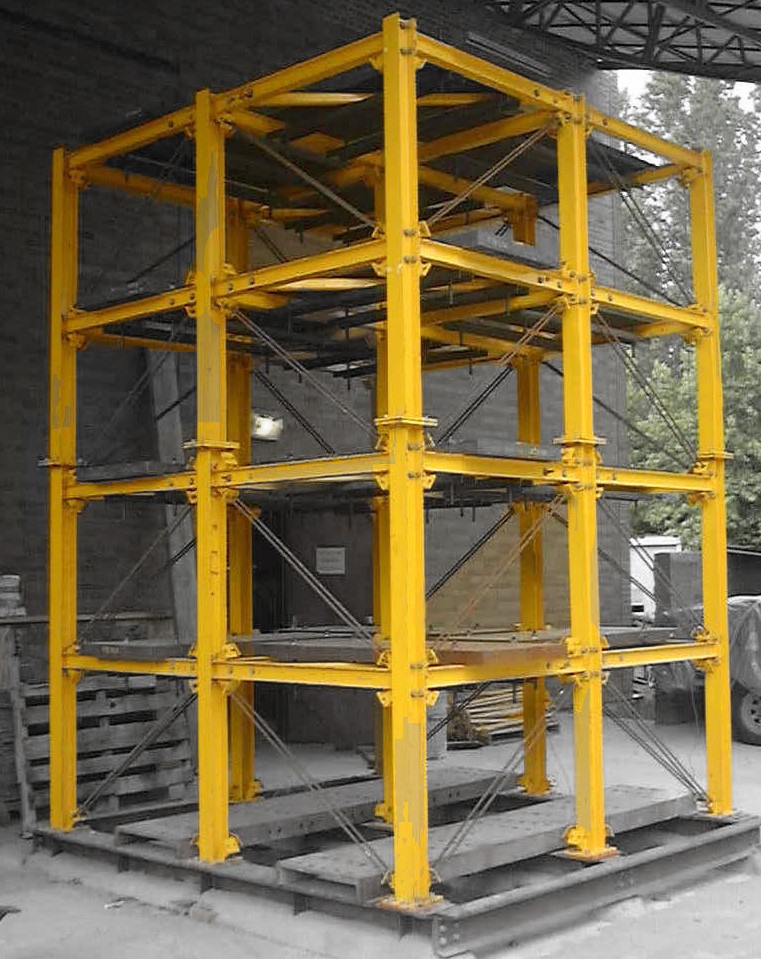}
        \caption {}
        \label{fig:Y1S}
    \end{subfigure}
    \begin{subfigure}[b]{0.48\linewidth}
        \centering
       \includegraphics[width=\textwidth]{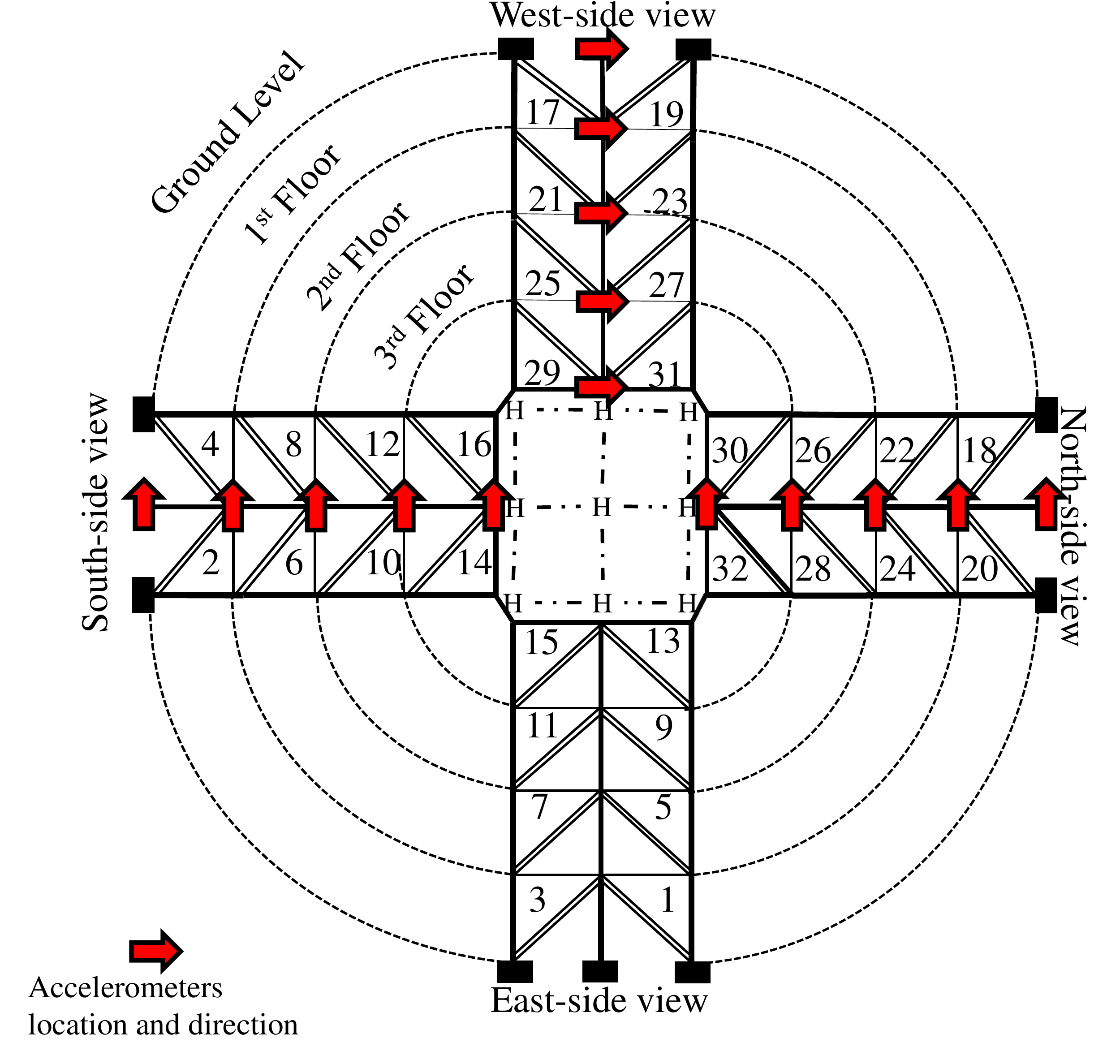}
        \caption{}
        \label{fig:Y2S}
    \end{subfigure}
    \caption{Yellow Frame (a) structure (in color) (b) brace labeling and sensors locations.}
    %\label{fig:approach1}
\end{figure}

Yellow Frame \cite{mendler2019yellow} is a one-third modular four-story steel frame established at the University of British Columbia, as shown in (Fig. \ref{fig:Y1S}). Modular frame components (i.e., braces, beams, and masses) can be removed to simulate diverse damage scenarios. There are fifteen accelerometers on the structure, capturing the ambient vibration response with a sampling frequency of 1000 Hz. Readers are referred to Mendler \textit{et al.} \cite{mendler2019yellow} and Allahdadian \cite{allahdadian2017robust} for details of the structure and instrumentation.

For this study, a benchmark dataset with twenty-one different structural configurations is selected, as presented in table \ref{W21}. This dataset includes one ``normal'' class and twenty damage classes, simulated by removing different structure braces. For creating time-series data objects from the captured acceleration data points, a vector of 1000 data points ($D_L$) is selected, resulting in the corresponding ``number of data objects'' also shown in the table.

\begin{table}[h!]
\begin{threeparttable}
  \caption{The dataset summary for Yellow Frame, including various damage classes simulated by removing certain bracing components, with labels shown in Fig. \ref{fig:Y2S}.}
  \label{W21}
  \centering
  \begin{tabular*}{\linewidth}{@{\extracolsep{\fill}} lccccc}
    \toprule
\parbox{ 0.8cm}{Damage class}&\parbox{3.5cm}{\centering Removed brace ID \\ (NO.)} &  \parbox{0.8cm}{\centering length (sec)}  & \parbox{0.8cm}{\centering Damage class}&\parbox{4cm}{\centering Removed brace ID \\ (NO.)} &  \parbox{0.8cm}{\centering  length (sec)} \\
    \midrule
FDC1&	None&	\parbox{0.8cm}{\centering 1047}&	FDC12&\parbox{4cm}{\centering FDC11 + (17, 19, 25, 27)(I)\tnote{*}}&	\parbox{ 0.8cm}{\centering 830}\\

FDC2&	2,4(II)\tnote{**}&	\parbox{ 0.8cm}{\centering 1147}&	FDC13&\parbox{4cm}{\centering FDC8 + (1, 3, 17, 18)(I)}&	\parbox{ 0.8cm}{\centering 1002} \\

FDC3&	FDC2 + (18, 20)(II)&	\parbox{ 0.8cm}{\centering 1147}&	FDC14&	(10, 12)(II)&	\parbox{ 0.8cm}{\centering 830}\\

FDC4&	\parbox{3.5cm}{\centering  FDC3+ (1, 3, 17, 19)(II)}& 	\parbox{ 0.8cm}{\centering 666}&FDC15	&\parbox{4cm}{\centering FDC14 + 21(II),  23(I)}&	\parbox{ 0.8cm}{\centering 1101}\\

FDC5&	FDC2 + (17, 19)(II)&	\parbox{ 0.8cm}{\centering 559}&	FDC16	&(21, 23)(II)&	\parbox{ 0.8cm}{\centering 713}\\

FDC6&	FDC2 + (18, 20)(I)&	\parbox{ 0.8cm}{\centering 665}&	FDC17&	\parbox{4cm}{\centering (7-8, 21, 22)(I)}&\parbox{ 0.8cm}{\centering 1050}\\

FDC7	&2(II)	&\parbox{ 0.8cm}{\centering 559}&	FDC18	&\parbox{4cm}{\centering (5, 6, 7, 8, 21, 24)(I)}&	\parbox{ 0.8cm}{\centering 1170}\\

FDC8&	(2 , 4)(I)&	\parbox{ 0.8cm}{\centering 659}&	FDC19	&\parbox{4cm}{\centering FDC18 + (7, 8, 21, 22)(I)}	&\parbox{ 0.8cm}{\centering 1170}\\

FDC9&	(25, 27)(I)&	\parbox{ 0.8cm}{\centering 668}&	FDC20	&\parbox{4cm}{\centering FDC19 + (5, 6,  23, 24)(I)}&	\parbox{ 0.8cm}{\centering 1003}\\

FDC10	&\parbox{3.5cm}{\centering (29, 31, 8, 6)(I)}&	\parbox{ 0.8cm}{\centering 1100}&FDC21&\parbox{4cm}{\centering	(6, 8)(II), (21, 22, 23, 24)(I)}&	\parbox{ 0.8cm}{\centering 1053}\\

FDC11	&\parbox{3.5cm}{\centering (21, 23 ,29 , 31)(I)}&	\parbox{ 0.8cm}{\centering 1000}&&&\\
\bottomrule
\end{tabular*}
\begin{tablenotes}\footnotesize
\item[*] One brace is removed
\item[**] Both braces are removed
\end{tablenotes}
\end{threeparttable}
\end{table}
%\phantom{A}
%\begin{figure}[h!]
  %  \centering
   %  \includegraphics[width=0.8\textwidth]{figures/Datasa%mpleYellow.eps}\\
   % \caption{Yellow Frame, class 14}
  %  \label{DD}
%\end{figure}
\subsection{Z24 Bridge}
 
\begin{figure}[h!]
    \centering
     \includegraphics[scale=.75]{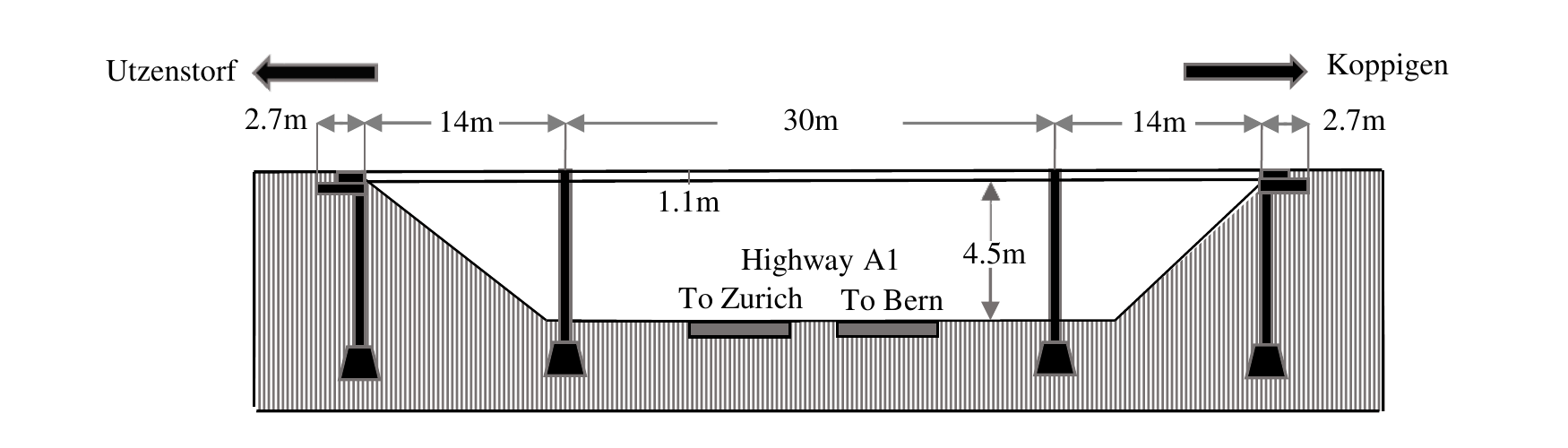}
    \caption{Z24 Bridge
            \label{Z24AA}
            \cite{reynders2009continuous}.}
    \end{figure}

Z24 Bridge overpassed the A1 highway between Bern and Zürich in Switzerland. A schematic of the bridge is shown in Fig. \ref{Z24AA}. This study's Z24 Bridge dataset consists of vibration and environmental monitoring of the bridge with various exerted damages before its demolition. Further information on the bridge and the sensor setups can be found in Reynders and De Roeck \cite{reynders2014vibration}. Based on a sampling rate of 100 Hz, a $D_L$ equal to 200 is chosen for making time-series data objects. The dataset summary, including different types of damages, is shown in Table \ref{Z24-D}. Among the 17 data classes, the first and the third reference measurements are not included in the detection process, as only one reference measurement is required for evaluating a novelty detection algorithm. The monitoring would be from the second reference measurement to the last tendon rupture damage (Table \ref{Z24-D}). 

The original dataset contained sixteen accelerometers with different orientations (i.e., vertical, horizontal, or transverse) on the bridge resulting in numerous data channels. To ensure that anomaly detection was limited to structural damages and not instrumentation failures (e.g., dead or square wave channels), we manually reduced the number of channels to those with no observable defects during the fifteen data classes, resulting in a total of eighteen channels. The selected channels are shown in Table \ref{Z24-A}, with the sensors' labels explained in the Z24 dataset \cite{reynders2009continuous}.

\renewcommand{\arraystretch}{1}
\begin{table}[h!]
  \caption{The dataset summary for Z24 Bridge, including various global and local damage scenarios.}
  \label{Z24-D}
  \centering
  \begin{tabular*}{\linewidth}{@{\extracolsep{\fill}} lcc}
    \toprule
\parbox{2cm}{Data class}&\parbox{5cm}{\centering Description \cite{reynders2014vibration}} &   \parbox{2.5cm}{\centering  Number of data objects} \\
    \midrule
BDC1&First reference measurement&325 (Not used) \\
BDC2&Second reference measurement ($Ref_{2}$)& 325\\
BDC3&Lowering of pier, 20 mm ($PL_{20}$)& 325 \\
BDC4&Lowering of pier, 40 mm ($PL_{40}$)&325 \\
BDC5&Lowering of pier, 80 mm ($PL_{80}$)&325 \\
BDC6&Lowering of pier, 95 mm ($PL_{95}$)& 325\\
BDC7&Lifting of pier, tilt of foundation ($T_{F}$)&325 \\
BDC8&Third reference measurement ($Ref_{3}$)&325 (Skipped) \\
BDC9&Spalling of concrete at soffit, $12 m^2$ ($SC_{12}$)& 325\\
BDC10&Spalling of concrete at soffit, $24 m^2$ ($SC_{24}$)&325 \\
BDC11&	Landslide of 1 m at abutment ($LS$)&325 \\
BDC12&Failure of concrete hinge ($F_{CH}$)&325 \\
BDC13&Failure of 2 anchor heads ($F_{2A}$)& 325\\
BDC14&Failure of 4 anchor heads ($F_{4A}$)&325 \\
BDC15&Rupture of 2 out of 16 tendons ($TR_{2}$)&325 \\
BDC16&Rupture of 4 out of 16 tendons ($TR_{4}$)&325 \\
BDC17&Rupture of 6 out of 16 tendons ($TR_{6}$)&325 \\
\bottomrule
\end{tabular*}
\end{table}
\phantom{A}
\renewcommand{\arraystretch}{1}
\begin{table}[h!]
  \caption{Selected channels' labels, Z24 Bridge dataset.}
  \label{Z24-A}
  \centering
  \begin{tabular*}{\linewidth}{@{\extracolsep{\fill}} l ccccc}
    \toprule
\parbox{1.5cm}{Label \cite{reynders2014vibration}}&\parbox{2cm}{\centering Sensor}&\parbox{2cm}{\centering Label \cite{reynders2014vibration}}&\parbox{2cm}{\centering Sensor}&\parbox{2cm}{\centering Label \cite{reynders2014vibration}}&\parbox{2cm}{\centering Sensor}\\
\midrule
Setup 1&R2V&Setup 2&106V&Setup 2&304V\\
Setup 2&R1V&Setup 3&R1V&Setup 4&116V\\
Setup 4 & 314V &Setup 4 & R2V  &Setup 5&  121V \\
Setup 5 & 319V &Setup 5&  R1V  &Setup 5 & R2V  \\
Setup 6&  126V &Setup 6&  R2V  &Setup 7&  131V \\
Setup 8&  136V &Setup 8&  R2V  &Setup 9&  R1V  \\
\bottomrule
\end{tabular*}
\end{table}

It is worth mentioning that on both datasets, the proposed detection algorithm starts from the reference class (i.e., BDC1, FDC1) and then continues to damaged classes in the same order that those classes were captured (e.g., FDC2 after FDC1). Having the same order of data in the data acquisition and novelty detection could result in unbiased detection results.

\section{Methodology}

\begin{figure}[h!]
    \centering
     \includegraphics[scale=.75]{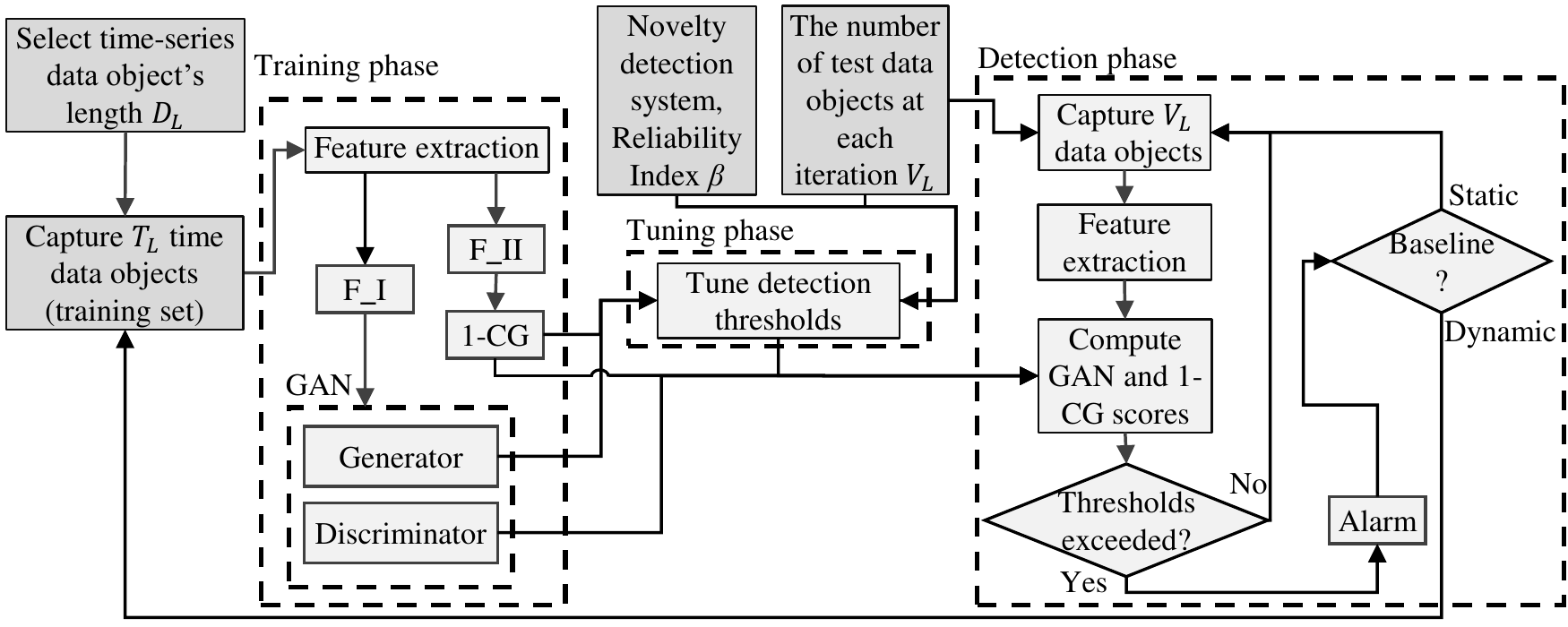}
 \caption {Unsupervised real-time SHM framework. Shaded blocks represent user-defined parameters.}
 \label{M1}
    \end{figure}

A general overview of the proposed unsupervised real-time SHM methodology is depicted in Fig. \ref{M1}. The method consists of three main phases. The training phase captures the first $T_L$ time-series data objects and extracts those data objects features for training the GAN and 1-CG models. It is assumed that all $T_L$ data objects belong to the same class. The trained GAN and 1-CG models output a probability for new data objects further used for novelty detection.
 
In the tuning phase, a pre-defined novelty detection system of limit-state functions' detection thresholds are dynamically tuned based on the user-defined parameters (i.e., $T_L$, $D_L$, and $V_L$) with reliability analysis. The tuning is vital for any method since non-proper parameters can lead to many false alarms, or undetected novelties \cite{bouzenad2019semi, de2019automated, santos2015baseline}. The Gan-generated data objects are utilized for the tuning phase to decrease the detection's sensitivity to user-defined parameters. Finally, the detection phase tests new data for novelties by either a static or dynamic baseline approach. In the static baseline approach, the detection thresholds are tuned only once, while in the dynamic baseline approach, the thresholds are tuned each time a novelty is detected. Further details are provided in the following subsections.

\subsection{Parameter Selection}

The novelty detection system requires four parameters to be defined by the user: $D_L$, $T_L$, $V_L$, and $\beta$. $D_L$ is the length of each time-series data object (the number of data points). The choice of $D_L$ is related to the structure under study and is defined to have a sensible frequency resolution and have a high-dimensional feature. $D_L$ for Yellow Frame and Z24 Bridge is defined as 1,000 (i.e., one second) and 200 (i.e., two seconds) data points, respectively.
 
The parameter $T_L$ (Fig. \ref{M1}) is the number of time-series data objects included in the training set. For generic novelty detection of a structure or across multiple structures, $T_L$ can be selected based on the monitoring duration, which can be several days or months. It is possible to use a dynamic $T_L$, in which the novelty detection framework is initiated with a short $T_L$, and it dynamically increased after a pre-defined period of novelty-free monitoring. For the two datasets studied herein, we target a value of $T_L$ resulting in a small percentage of each class's available time-series data objects (i.e., high test-to-train ratio). A high test-to-train ratio creates challenges in the algorithm's sensitivity to false alarms, one of our critical criteria for evaluating the framework's performance. It can also illustrate the algorithm's sensitivity to novelties (true alarms) with limited training data.
 
For Yellow Frame, we selected $T_L=100$, allocating between 10\% and 16\% of data objects to the training set (i.e., test-to-train ratio of 5 to 9). For the Z24 dataset, a $T_L=75$ results in 23\% of data objects being included in the training set. Both test-to-train ratios are much less than the conventional test-to-train ratio of 0.25 (80\% train-validation, and 20\% test). The parameter $V_L$ is defined as the number of time-series data objects monitored at each iteration of the detection phase. The proposed novelty detection framework tries to tune the novelty detection approach to reduce its sensitivity to the user's choice of $V_L$.

%AAA
\subsection{Training Phase}

GAN and 1-CG models are trained with the features extracted from the first $T_L$ time-series data objects in the training phase. In this section, first, the feature extraction method is explained. Then, GAN's architecture and 1-CG model are defined. Furthermore, two scores are determined based on the trained GAN and 1-CG models for further utilization in the tuning and detection phases.

\subsubsection{Feature Extraction}

\begin{figure}[h!]
    \centering
     \includegraphics[scale=.8]{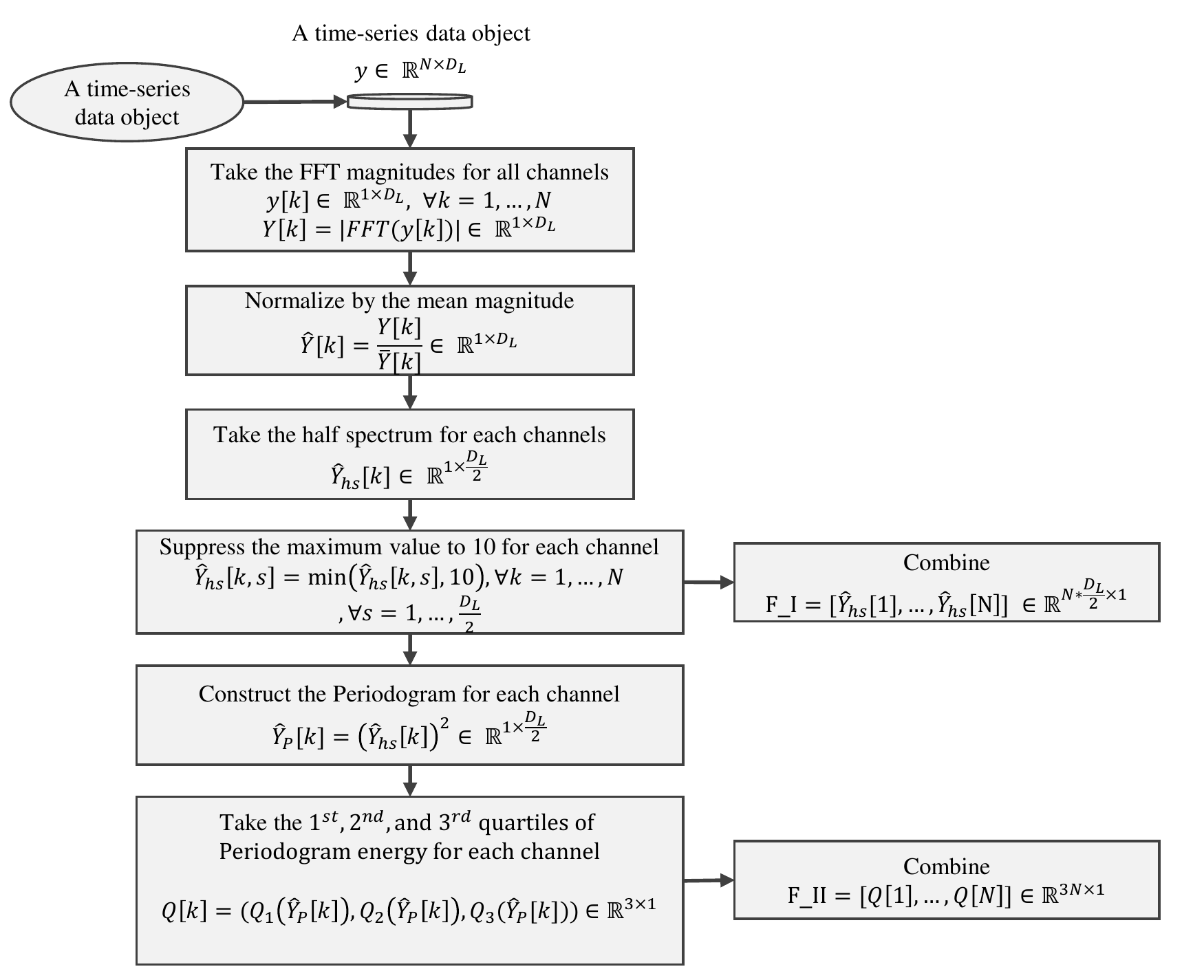}
 \caption {Feature extraction step, $N$ is the number of data channels.}
 \label{FE}
    \end{figure}

Based on the difficulties caused by the \textit{loss of information} and the \textit{case-dependency} in real-time SHM, we implement a simple feature extraction method that balances high- and low-dimensional features, as shown in Fig. \ref{FE}. The extraction is based on the fast Fourier transform (FFT) of the input signals. The first feature (F\_I) is high-dimensional ($N\times D_L/2$), made of magnitudes of half-spectrum FFTs of input signals with maximum values suppressed to ten. As discussed in Soleimani \textit{et al.} \cite{soleiman2020}, the features to vary in a fixed-range is beneficial for the GAN's training. The second feature (F\_II) is a reduced F\_I representation, made of quartiles of vibrational energy in each time-series data object, which is tried on prior studies \cite{de2019automated}. Since the method is designed to be general and avoid time-averaging with specified windows, a Periodogram power spectral density estimation is employed. In this method, FFT magnitudes of a signal are normalized and raised to the power of two to build the estimation. The estimation can be used to extract vibrational energy quartiles. Both low- and high-dimensional features are utilized in the detection phase, and their effectiveness is discussed.

\subsubsection{The GAN architecture}

\begin{figure}[h!]
    \centering
     \includegraphics[scale=.75]{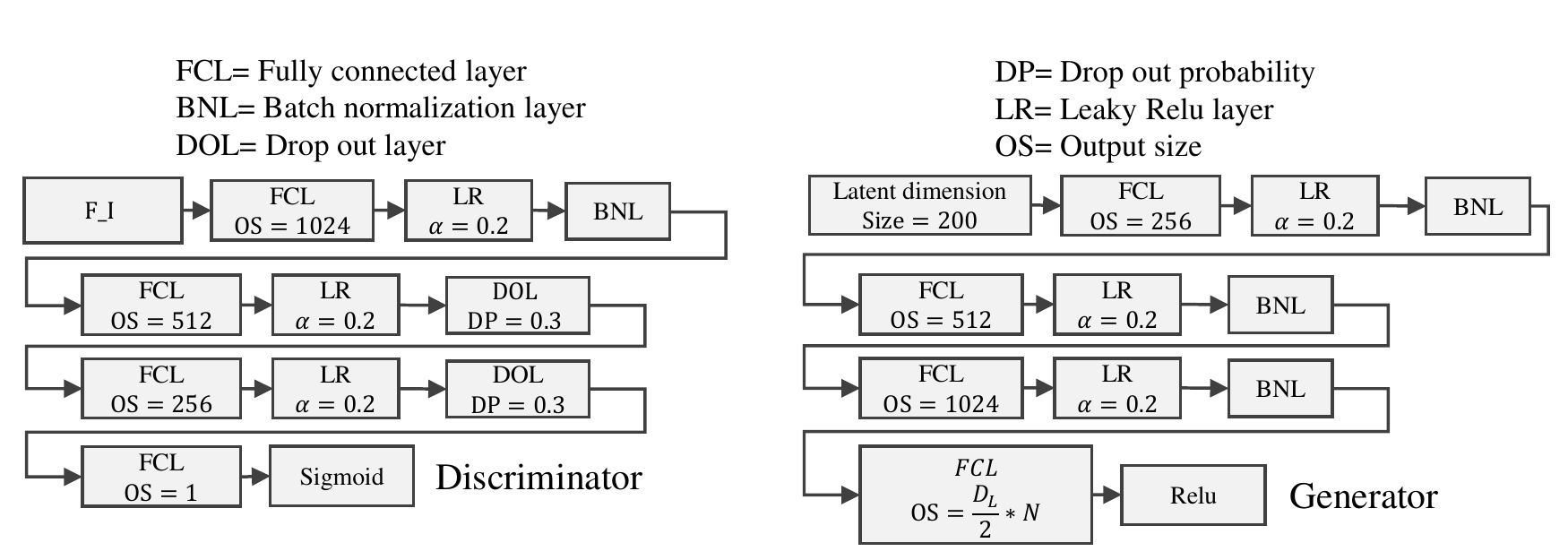}
 \caption {The GAN architecture.}
 \label{GANARC}
    \end{figure}

GAN consists of two neural networks, a generator (Generator) and a discriminator (Discriminator). The Generator attempts to produce a fake set of F\_I features. In contrast, the Discriminator distinguishes them from real ones (i.e., training set), treating the problem as a binary classification. The competition between the Generator and the Discriminator results in a Discriminator understanding the non-trivial patterns in the data objects' F\_I features. In turn, the Generator learns to generate random F\_I features that appear realistic; that is, they mimic the training set's F\_I features. In our methodology, the Generator is used to generate new sample features in the tuning phase, while the Discriminator is used as a binary classifier in the detection phase. The GAN's architecture is defined as shown in Fig. \ref{GANARC}. The GAN's Generator is fed by a vector of random numbers drawn from a standard Gaussian distribution (a latent space) with a size of 200. It is worth mentioning that in both datasets, $\frac{D_L}{2}\times N$ value is greater than the number of neurons in the penultimate layer of the Generator, and users can change this layer's neurons to meet their problem.

There is no specific criterion in the GAN's training to specify when the training can be stopped. A user can evaluate the GAN's training based on training-loss diagrams to conclude its training; however, we chose a static number of epochs in this study. For Yellow Frame, the F\_I feature has a dimension of $7500\times1$, and the number of epochs is set to be 5000. For the Z24, the F\_I feature has a dimension of $1800\times1$, and the epoch number is set to be 2000. The training batch size is equal to $T_L$ (i.e., the entire training set); hence, each epoch is equivalent to an iteration. The Adam optimizer \cite{kingma2014adam} is used with a learning rate, $\beta_1$, and $\beta_2$ values equal to 0.0002, 0.5, and 0.999, respectively. 
 
The trained Generator's output is an $N* D_L/2 \times 1$ vector (i.e., the F\_I feature), which mimics the $T_L$ training objects' behavior when given an input vector of 200 random values. The Discriminator's output to a F\_I feature, $O_{dis}(F_I)$, is the output of a sigmoid activation layer with a value between zero and one. The higher this value, the higher the Discriminator's confidence that a new test data object belongs to GAN's training set or not. From this, we define the GAN's Discriminator-based score, $S_{GAN}$ for a test data object's F\_I feature, as 
\begin{equation}
\label{EQ1}
    S_{GAN}(\textrm{F\_I})=-\textrm{log}_{10}({O_{dis}(\textrm{F\_I})).}
\end{equation}
Similarly to the Discriminator's output, $S_{GAN}$ expresses the confidence that the features come from the training set class but using a log scale to increase the scale of numeric outputs.

\subsubsection{1-CG architecture}

The 1-CG classifier is a joint Gaussian distribution with a dimension of $3N$. This distribution is fitted with an expectation-maximization algorithm to the training F\_II features. It is used to compute the score $S_{1-CG}$ for a new test data object as defined in Eq. \ref{EQ2}, in which $NL$ represents the negative log-likelihood from the trained joint Gaussian distribution. In a density-based anomaly detection manner, the $NL$ of a test data object's F\_II feature is divided by the mean $NL$ values of the training set's F\_II features (F\_II\_T) to detect anomalies based on the resulting score.
\begin{equation}
\label{EQ2}
    S_{1-CG}(\textrm{F\_II})=\frac{\textrm{NL}_{1-CG(\textrm{\textrm{F\_II}})}}{\textrm{mean}(\textrm{NL}_{1-CG(\textrm{\textrm{F\_II\_T}})})}.
\end{equation}
\subsection{Tuning and detection phase}

The trained GAN and 1-CG models produce two different scores, namely $S_{GAN}$, and $S_{1-CG}$, both of which examine the chance of a test data object to be an outlier (i.e., novelty). In the tuning phase, a pre-defined novelty detection system of limit-state functions is tuned to the user-defined parameters (i.e., $D_L$, $T_L$, and $V_L$). The novelty-detection system and the proper methods for tuning it are explained in further subsections.
\subsubsection{The novelty detection system of GAN and 1-CG ensemble}
\begin{figure}[h!]
    \centering
    \begin{subfigure}[b]{0.48\linewidth} 
        \centering
     \includegraphics[width=\textwidth]{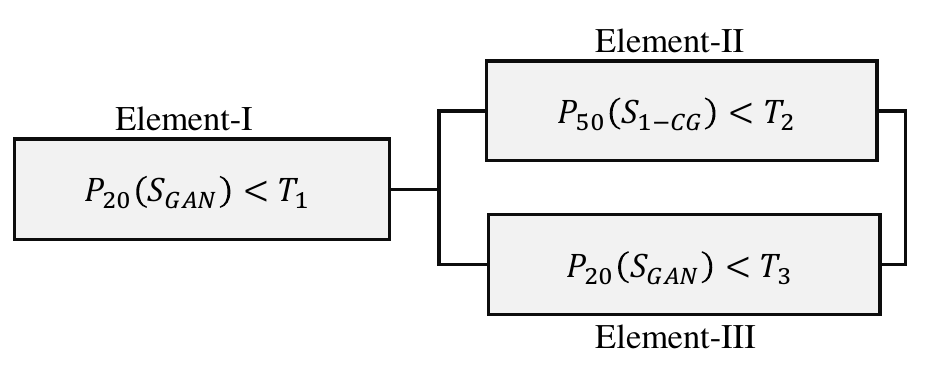}
        \caption {}
        \label{fig:Y1}
    \end{subfigure}
    \begin{subfigure}[b]{0.48\linewidth}
        \centering
       \includegraphics[width=\textwidth]{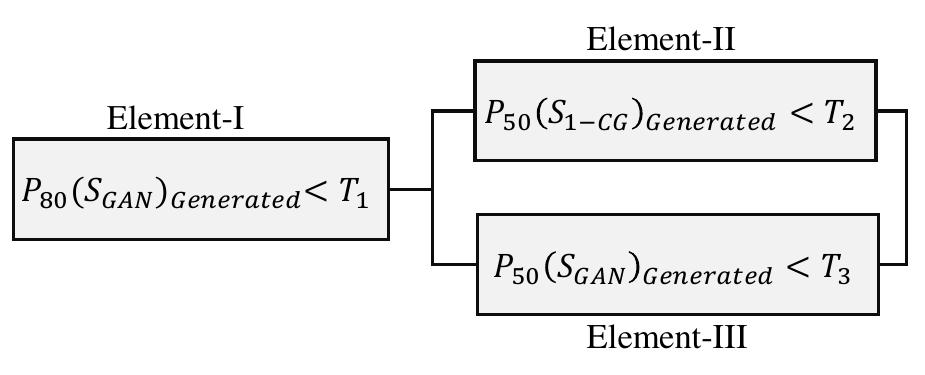}
        \caption{}
        \label{fig:Y2}
    \end{subfigure}
    \caption{Novelty detection systems for (a) unseen data (b) tuning thresholds against GAN-generated data objects.}
    %\label{fig:approach1}
\end{figure}

Generally, a limit-state function ($g$) can be defined as
\begin{equation}
\label{Rg}
    g(T,S)=T-S,
\end{equation}
which can be viewed as the difference between Resistance ($T$) and Load ($S$). A limit-state function fails whenever $S$ is greater than $T$. Defining a combination of different limit-state functions in terms of $S_{GAN}$ and $S_{1-CG}$ for a test data object can result in a novelty detection system. The resultant system can be a combination of series and parallel limit-state functions (elements). It is desirable for $S_{GAN}$ be more important in the novelty detection system since it is based on a high-dimensional feature. Hence, the novelty detection system is defined as shown in Fig. \ref{fig:Y1}, in which the $S_{GAN}$-based score is in the majority (i.e., two out of three elements) and is an absolute stand (i.e., the serial element). There are three elements in the defined novelty detection system, each of which has its limit-state function. The thresholds (i.e., $T_1$, $T_2$, and $T_3$) can be viewed as $T$, and different percentiles of $S_{GAN}$ and $S_{1-CG}$, as $S$. The system detects a novelty if $S$ is over $T$ in the Element-I or both Elements II and III simultaneously fail. It is worth mentioning that Element-I and Element-III have the same $T$; hence, a more relaxed $S$ must be selected for Element-III compared to Element-I, as a parallel element.

As outlined in Fig. \ref{M1}, at each iteration of the detection phase, $V_L$ (i.e., detection-window length) of data objects are used to capture a $V_L\times 1$ vector of $S_{GAN}$ and $S_{1-CG}$. The $P20$, $P80$, and $P50$ are different percentiles (i.e., 20\textsuperscript{th}, 80\textsuperscript{th}, and 50\textsuperscript{th} percentiles) of the resulting vector of scores. The main idea behind selecting the percentiles of scores is to reduce the noisy scores' effect on novelty detection. The tunable parameters in the system are the thresholds: $T_{1}$, $T_{2}$, and $T_{3}$. Since this is an unsupervised approach, except for the $T_L$ training data objects, no more data is available to tune the thresholds. A key novelty in the approach is that it leverages the Generator-generated data objects to tune the thresholds to a specified sensitivity to novelties, regardless of the choice of other user-defined parameters. This is achieved by defining an analogous system shown in Fig. \ref{fig:Y2}. The analogous system must not fail (i.e., have a low probability of failure) for the GAN-generated data objects. This approach assumes that the variation in the GAN-generated data objects is similar to or greater than the variation of the unseen data belonging to the same class as the training set, thus representing extreme cases.

The Generator can only generate F\_I features, but F\_II features can be constructed from the F\_I features (Fig. \ref{FE}). The Generator-generated F\_I and F\_II features are initiated from a standard Gaussian distribution and are random variables. Hence, for a system of limit-state functions with random variables, applying a reliability-based analysis to tune the thresholds to the desired reliability index from the Generator-generated time-series data objects is possible. As mentioned, the reliability analysis must be carried out on the analogous system shown in Fig. \ref{fig:Y2}. In the main system, specific percentiles of the scores are used to avoid noisy detection scores. However, while tuning the system's thresholds with the Generator-generated data objects, the extreme scores must be considered. Thus, any $F$\textsuperscript{th} percentile lower than the 50\textsuperscript{th} percentile, in the detection system, is transformed into the $(100-F)$\textsuperscript{th} percentile in the analogous system. As discussed, since Elements I and III have the same left side, its threshold must be more relaxed in the analogous system than the main system. Hence, the 50\textsuperscript{th} percentile rather than the 80\textsuperscript{th} percentile is used for Load. Since the thresholds are the same for both systems, the analogous system's tuning avoids flagging an alarm for unseen data objects from the training class since it was tuned to more extreme Generator-generated data objects.

\subsubsection{The reliability analysis method and unreliable GAN-generated data objects}

As already mentioned, the thresholds must be selected to avoid damage detection on the Generator-generated data objects as much as possible. A reliability index ($\beta$) is an indicator of the failure probability of limit-state function(s). A higher $\beta$ means a lower failure probability. Since the analogous system's failure should have a very low probability, a high $\beta$ value is desirable. In this study, $\beta$ of 3 is chosen as a suitable value, which corresponds to a 0.0013 probability of failure for the analogous novelty detection system. Additional values of $\beta$ are tried once for a constant $V_L$ to demonstrate its effect on novelty detection in the results section. By assuming the novelty detection system's elements to be uncorrelated (for simplicity), the system's reliability can be summarized as
\begin{equation}
\label{R}
    R_{system}=R_{EL-I}(R_{EL-II}+R_{EL-III}-R_{EL-II}R_{EL-III})
\end{equation}
Furthermore, the system can be tuned by assigning the same reliability ($R_{E}$) to all three elements, resulting in
\begin{equation}
\label{R2}
    R_{system}=2{R_{E}}^2-{R_{E}}^3
\end{equation}
By solving Eq. \ref{R2} for the $\beta$ of 3, the resultant $\beta$ for each element is 3.012. For each element, Monte Carlo histogram sampling (MCHS) can be applied. In MCHS, $S$ of the analogous novelty detection system elements (i.e., percentiles of $S_{GAN}$, and $S_{1-CG}$) are sampled, and histograms of their values are constructed. Based on the acquired histogram, $T$ (i.e., thresholds) can be selected to satisfy the reliability index criterion. In this study, and based on the target reliability index, 5000 iterations are used in the MCHS; however, a higher $\beta$ would require more iterations. 

The Generator generates F\_I features from a standard Gaussian distribution (SGD), and numerous neurons, weights, and biases transform the random noise into realistic fake F\_I features. There is always the possibility that the Generator is not trained well for all parts of that SGD, resulting in misleading generations. This phenomenon is shown in Soleimani \textit{et al.} \cite{soleiman2020}. In that study, generated data objects were ``Capped'' to avoid misleading generations to enter the analysis. There should be a filter to perform the same task on the MCHS-generated histograms of Loads in this study. The following algorithm is defined based on a two-class Gaussian Mixture model (2-CGMM). The algorithm is set to reach an equilibrium state (i.e., detecting no more outliers) fast to avoid excess data cleaning by toughening the cleansing threshold in successive iterations (Alg. \ref{Alg1}). 

The proposed novelty detection system can be tuned for all user-defined parameters; however, the most influential parameter is $V_L$, as it dictates the length of the vector of time-series data objects to be monitored on each iteration. Hence, in this study, different values of $V_L$ are used for evaluating how the reliability-based tuning phase lowers the detection sensitivity to its selection. With a tuned novelty detection system, trained GAN, and 1-CG, the detection phase can be initiated. The detection phase can have a static baseline (i.e., one class) or a dynamic baseline. The only difference between the static and dynamic baseline approaches (Fig. \ref{M1}) is that in the static one, the thresholds are tuned only once for the normal class (i.e., the initial data stream). Damages are detected against the normal condition, without the potential to have discrimination amongst different damage scenarios. Dynamic baseline; however, tunes thresholds after each novelty is detected, understanding and preserving different damage scenarios' characteristics.

\begin{algorithm}[H]
\label{Alg1}
\SetAlgoLined
  \KwInput{A Load histogram}
  \KwOutput{An outlier-removed Load histogram}
$i\gets 1$\;
$j\gets 1$\;
 \While{j==1}{
  Train a 2-class GMM on the Load histogram\;
    \eIf{the second cluster mean value divided by the first cluster mean value is greater than i+1}{
   Delete the second cluster points\;
   $i\gets i+1$\;
   }{
 $j\gets 0$\;
  }
 }
 \caption{2-CGMM-based anomaly detection in score histograms.}
\end{algorithm}

\section{Yellow Frame results and discussion}

This study aims to define a framework in which the novelty detection thresholds are tuned based on the user-defined parameters so that false alarms and novelty detection's accuracy become uncorrelated to the chosen parameters. For Yellow Frame, $T_L$ is equal to one-hundred, and three values of ten, twenty, and forty are chosen for $V_L$, and the detection method is performed on the dataset. In what follows, results from each segment of the proposed unsupervised real-time SHM method (Fig. \ref{M1}) are presented and discussed. Furthermore, and only for the case of $V_L=100$, different $\beta$ values are employed in the tuning phase to evaluate its effect on the results.

\subsection{GAN and 1-CG training, Yellow Frame}

\begin{figure}[h!]
    \centering
    \begin{subfigure}[b]{0.48\linewidth} 
        \centering
     \includegraphics[width=\textwidth]{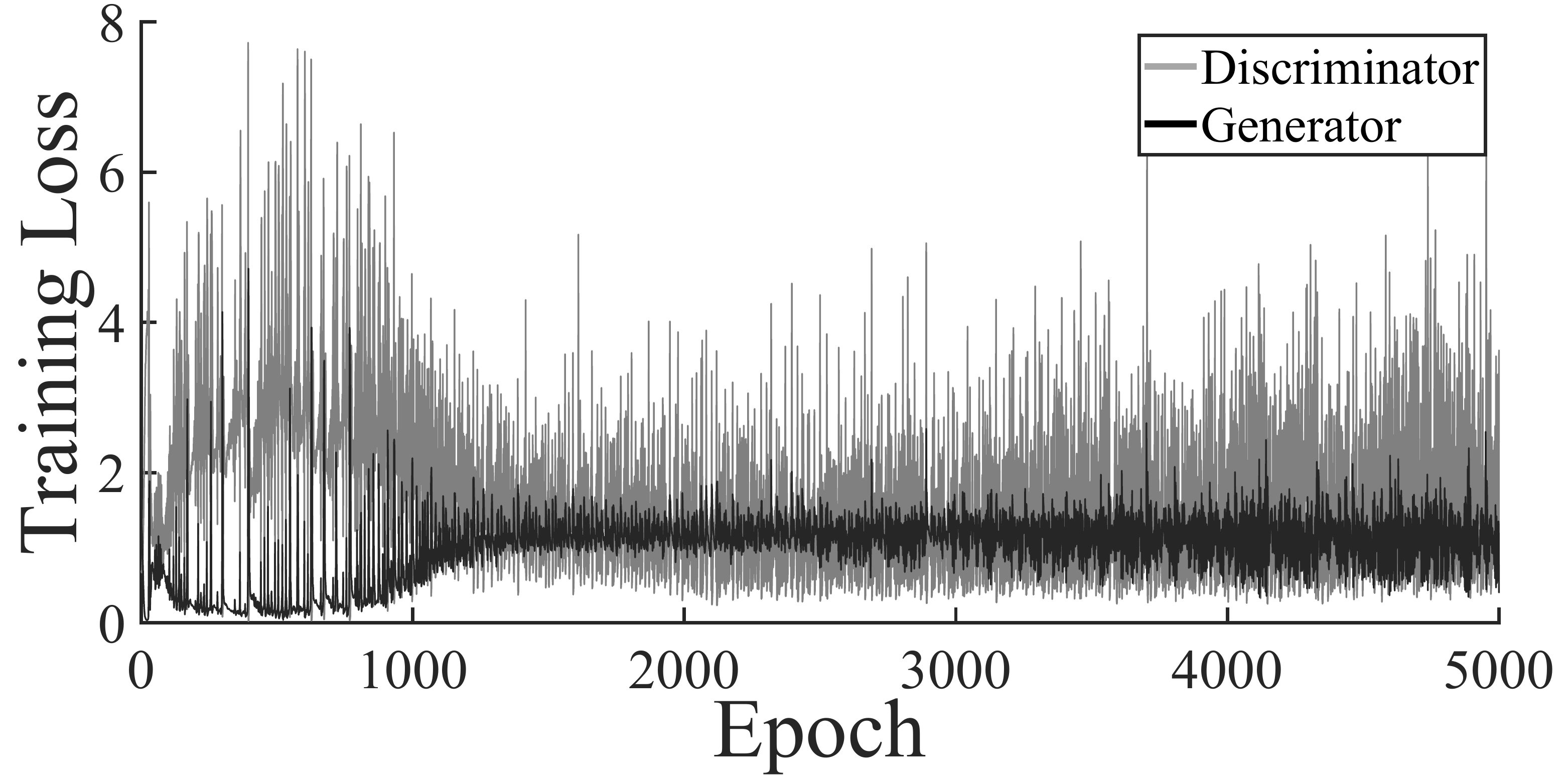}
        \caption {}
        \label{fig:GYT}
    \end{subfigure}
    \begin{subfigure}[b]{0.48\linewidth}
        \centering
       \includegraphics[width=\textwidth]{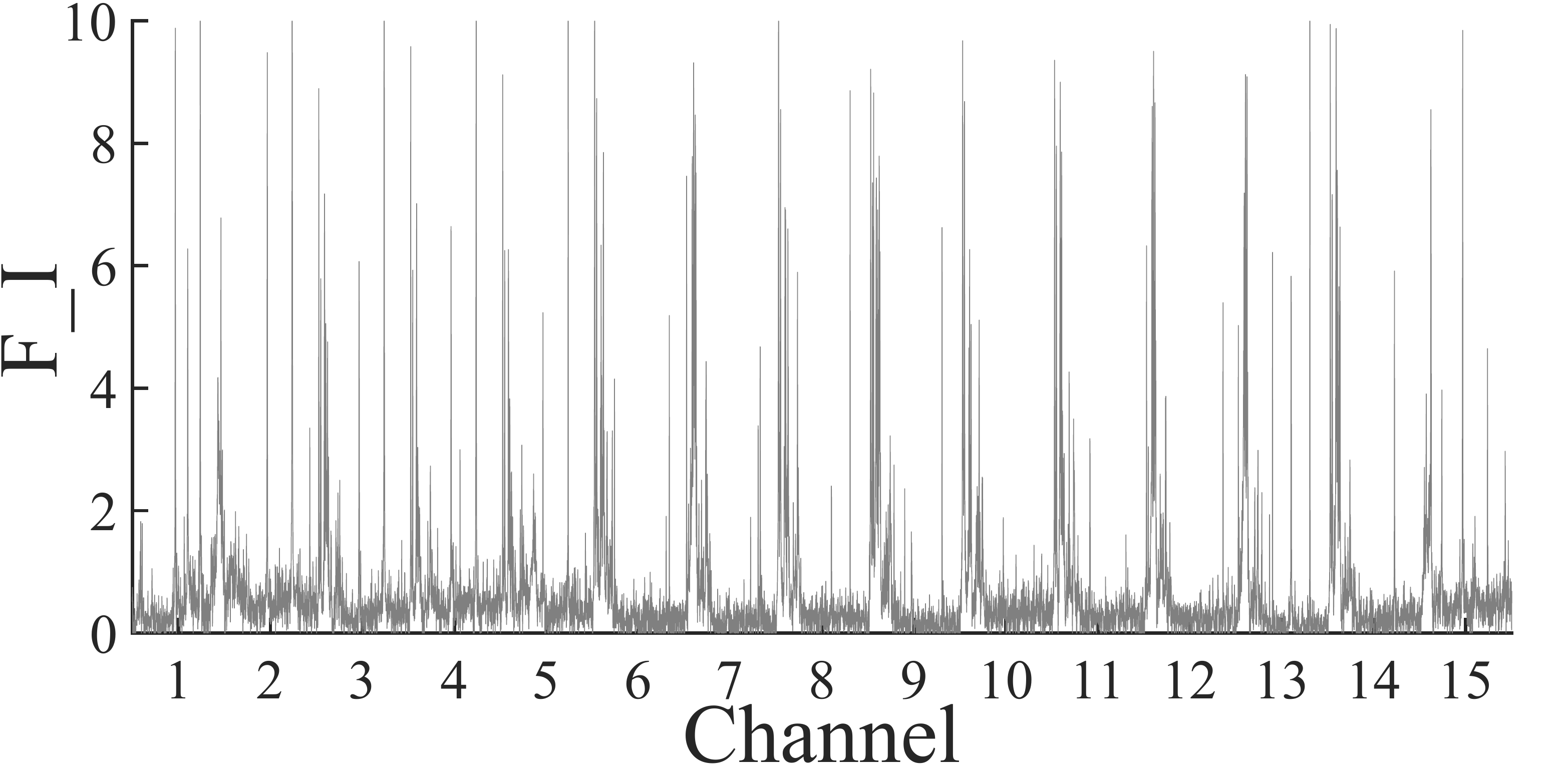}
        \caption{}
        \label{fig:GYT2}
    \end{subfigure}
    \caption{Unsupervised real-time SHM framework results for FDC1 (a) GAN' training loss (b) a generated F\_I feature.}
    %\label{fig:approach1}
\end{figure}

Although the GAN's training is performed with no supervision, metrics such as the Generator's training loss defined as 
\begin{equation}
 \label{SD}
     DL= -\textrm{mean}(\log{(OR+\epsilon)}+\log{(OF+\epsilon)})
 \end{equation}
and Discriminator's training loss defined as
\begin{equation}
 \label{SG}
     GL= -\textrm{mean}(\log{(OF+\epsilon)})
 \end{equation}
where $OF$ and $OR$ are Discriminator's outputs for the training set and the Generator-generated data objects (i.e., F\_I feature) are observable to decide on the GAN's training termination. The GAN's training loss and a generated F\_I feature for class FDC1 are shown in Fig. \ref{fig:GYT} and Fig. \ref{fig:GYT2}, respectively. The1-CG model is also trained on the training set (i.e., $T_L$ data objects).

\subsection{Novelty-detection system's tuning for $V_L$}

\begin{figure}[h!]
    \centering
    \begin{subfigure}[b]{0.32\linewidth} 
        \centering
     \includegraphics[width=0.975\textwidth]{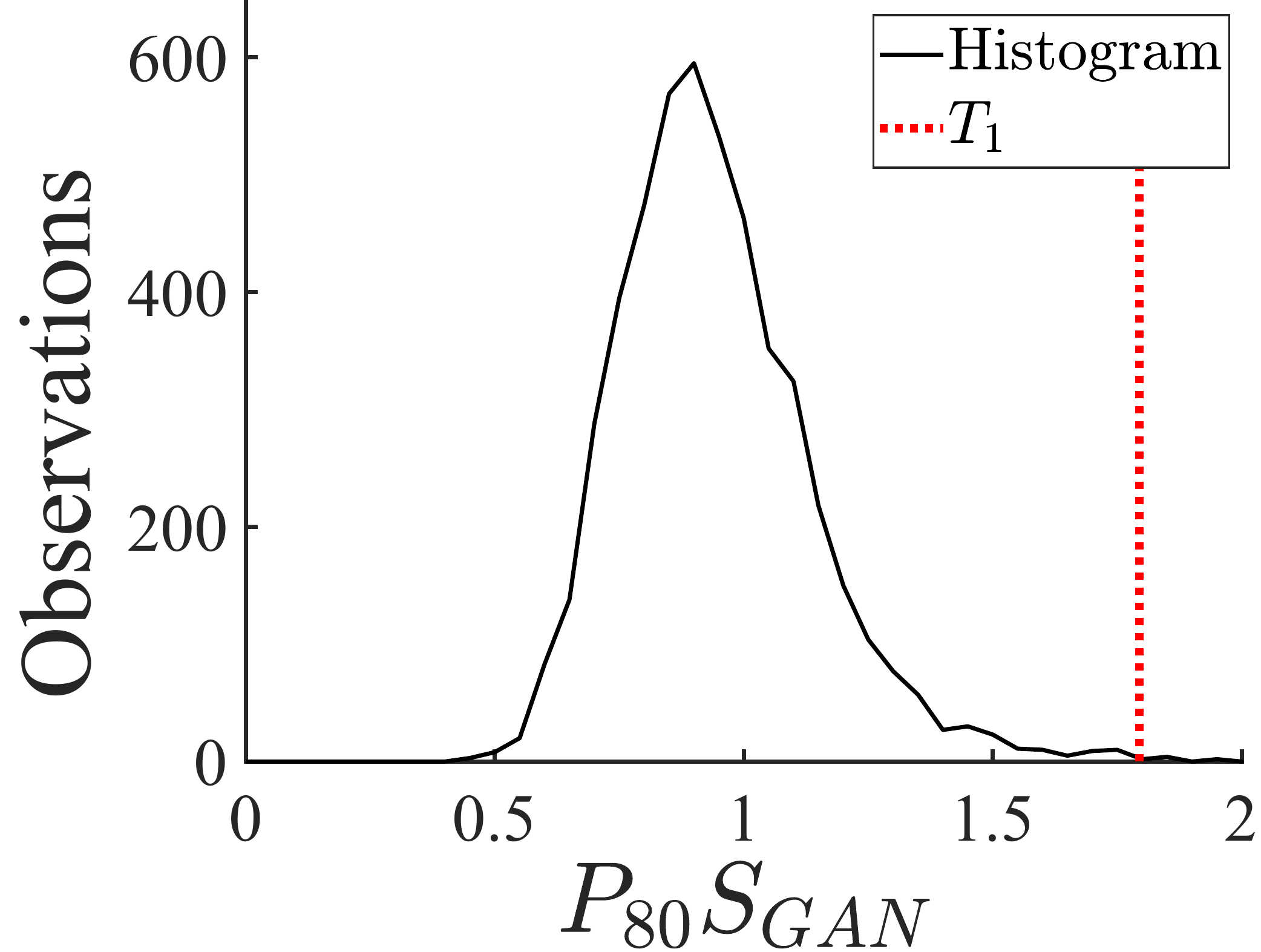}
        \caption {}
        \label{fig:GYO7}
    \end{subfigure}
    \begin{subfigure}[b]{0.32\linewidth}
        \centering
       \includegraphics[width=0.975\textwidth]{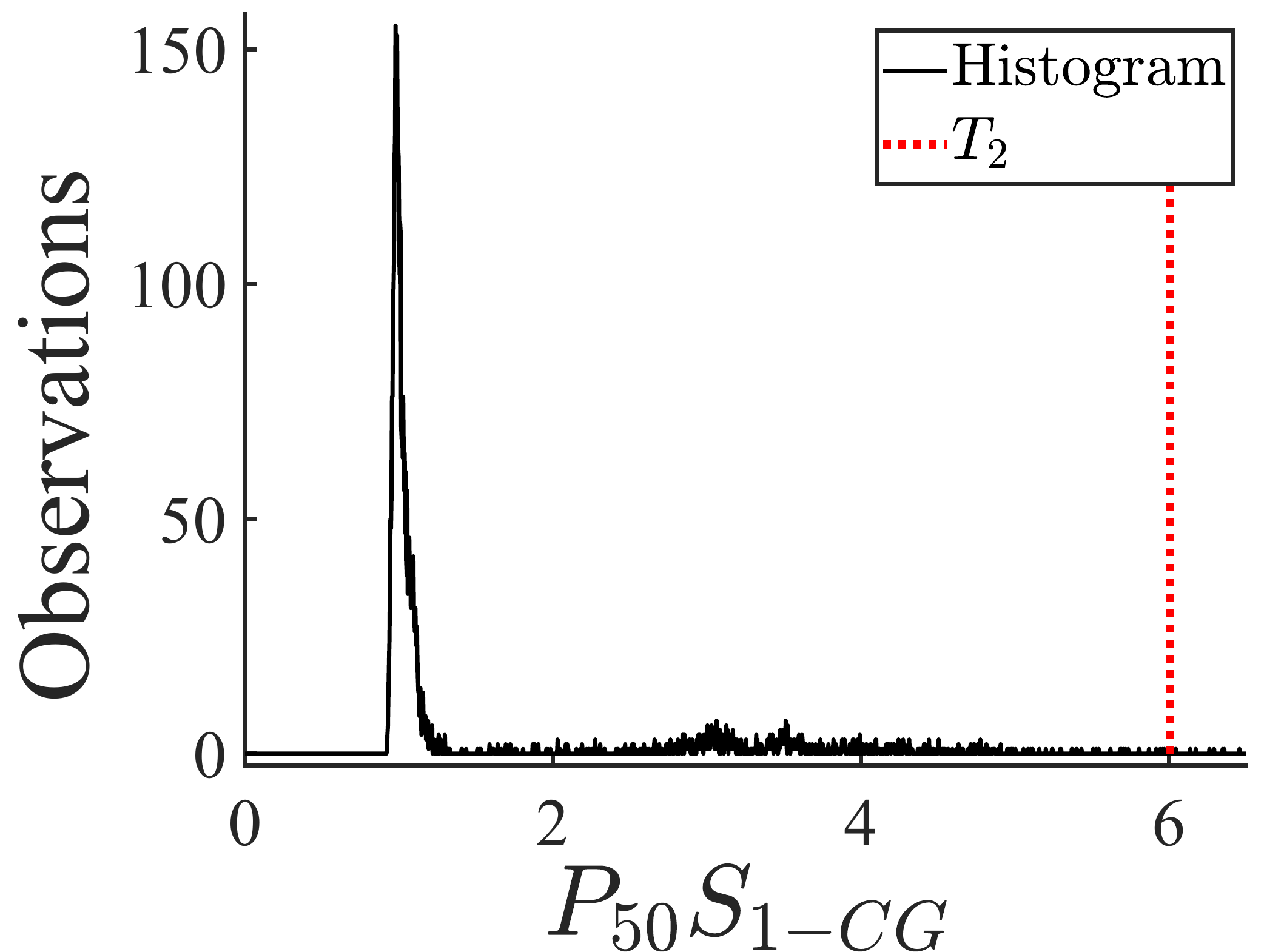}
        \caption{}
        \label{fig:GY2O5}
    \end{subfigure}
        \begin{subfigure}[b]{0.32\linewidth}
        \centering
       \includegraphics[width=0.975\textwidth]{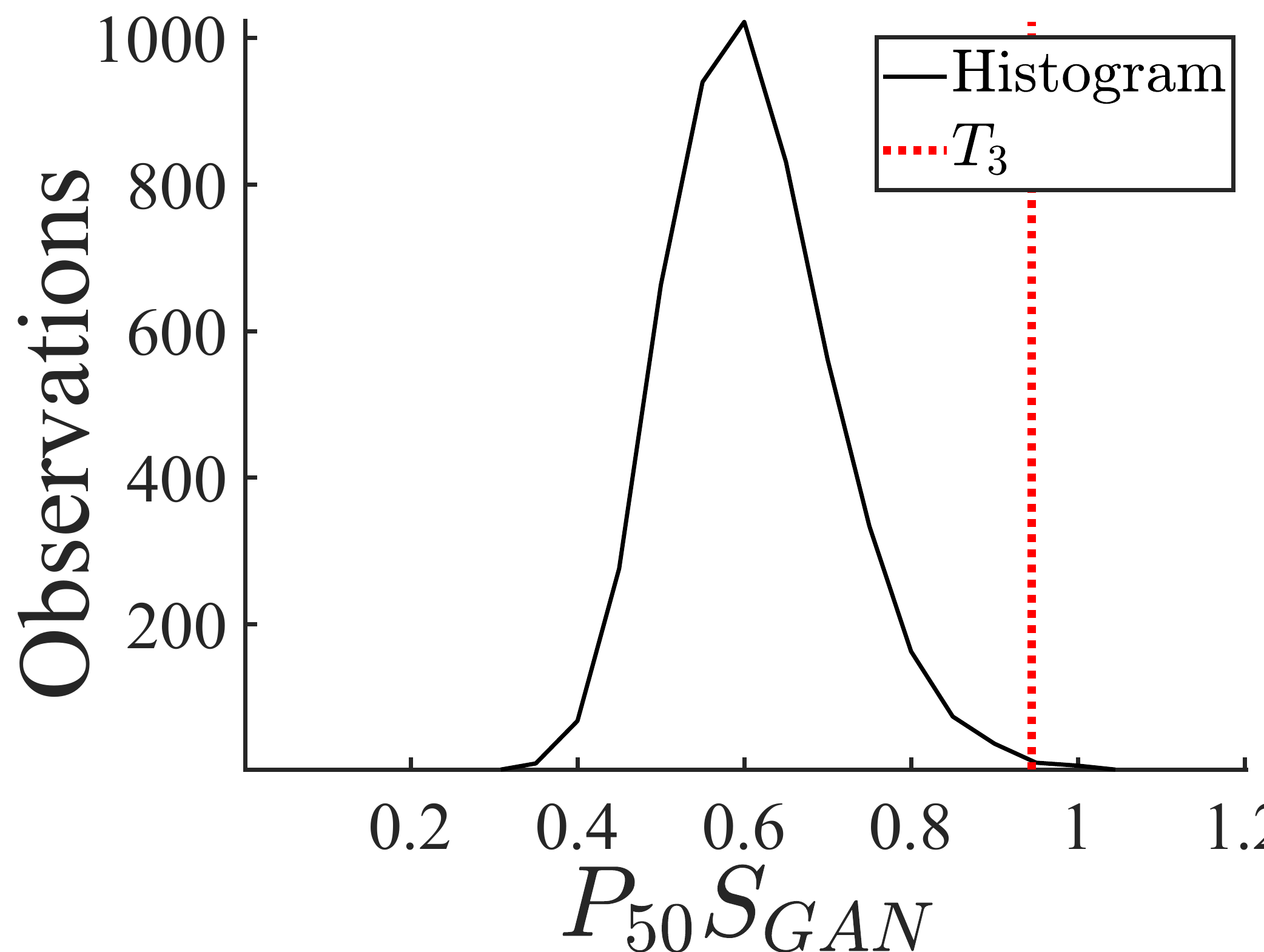}
        \caption{}
        \label{fig:GY2O6}
    \end{subfigure}
    \caption{Load MCHS results of novelty-detection system's Elements (a) I (b) II (c) III.}
    \label{BO}
    %\label{fig:approach1}
\end{figure}

\begin{figure}[h!]
    \centering
    \begin{subfigure}[b]{0.32\linewidth} 
        \centering
     \includegraphics[width=0.975\textwidth]{figures/F1.pdf}
        \caption {}
        \label{fig:GY}
    \end{subfigure}
    \begin{subfigure}[b]{0.32\linewidth}
        \centering
       \includegraphics[width=0.975\textwidth]{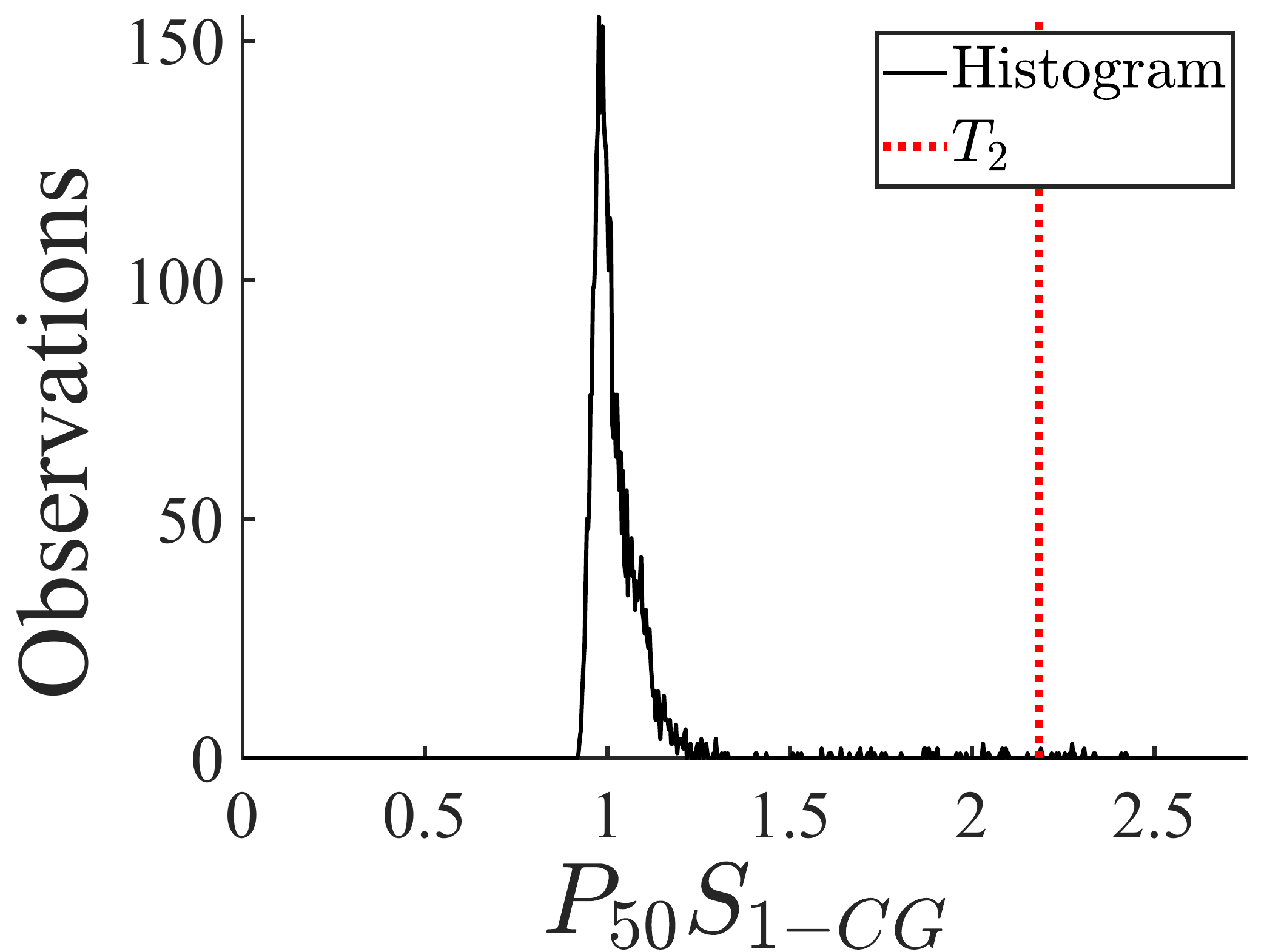}
        \caption{}
        \label{fig:GY2}
    \end{subfigure}
        \begin{subfigure}[b]{0.32\linewidth}
        \centering
       \includegraphics[width=0.975\textwidth]{figures/F3.pdf}
        \caption{}
        \label{fig:GY3}
    \end{subfigure}
    \caption{MCHS results (without outliers) of novelty-detection system's Elements (a) I (b) II (c) III.}
    \label{AO}
    %\label{fig:approach1}
\end{figure}

With the trained Generator, Discriminator, and 1-CG, MCHS of scores (i.e., Loads) in all three elements before and after the application of 2-CGMM (i.e., outlier-detector) for FDC1 is shown in Fig. \ref{BO}, and Fig. \ref{AO}, respectively. MCHS results exhibit a considerable tail for $P_{50}S_{1-CG}$. Since the reliability analysis seeks extreme values (i.e., high $\beta$), the resulting detection threshold would grow too large without the application of 2-CGMM, and no detection would occur. 2-CGMM implementation can result in no change in the Load histogram. The Element-I's and Element-III's Load histograms shown in Fig. \ref{BO}, and Fig. \ref{AO} show no change with the 2-CGMM.


\subsection{Unsupervised real-time novelty detection results, Yellow Frame}

With the determination of the novelty detection system's thresholds for the first class of data (i.e., FDC1), $V_L$ time-series data objects are analyzed using the novelty detection method each iteration. If a novelty is detected before the starting point of a new class of damage minus $V_L$, the detection is labeled as a false alarm; otherwise, the detection can be on-time or with delay, or novelties can be undetected. In the dynamic baseline approach, GAN and 1-CG training phases are performed after a novelty is detected, and this cycle continues for the entire dataset. For the sake of enumerating the false alarms, if a false alarm is triggered, the analysis continues until the detection of a  true novelty happens. 

\begin{figure}[h!]
    \centering
     \includegraphics[width=0.9\textwidth]{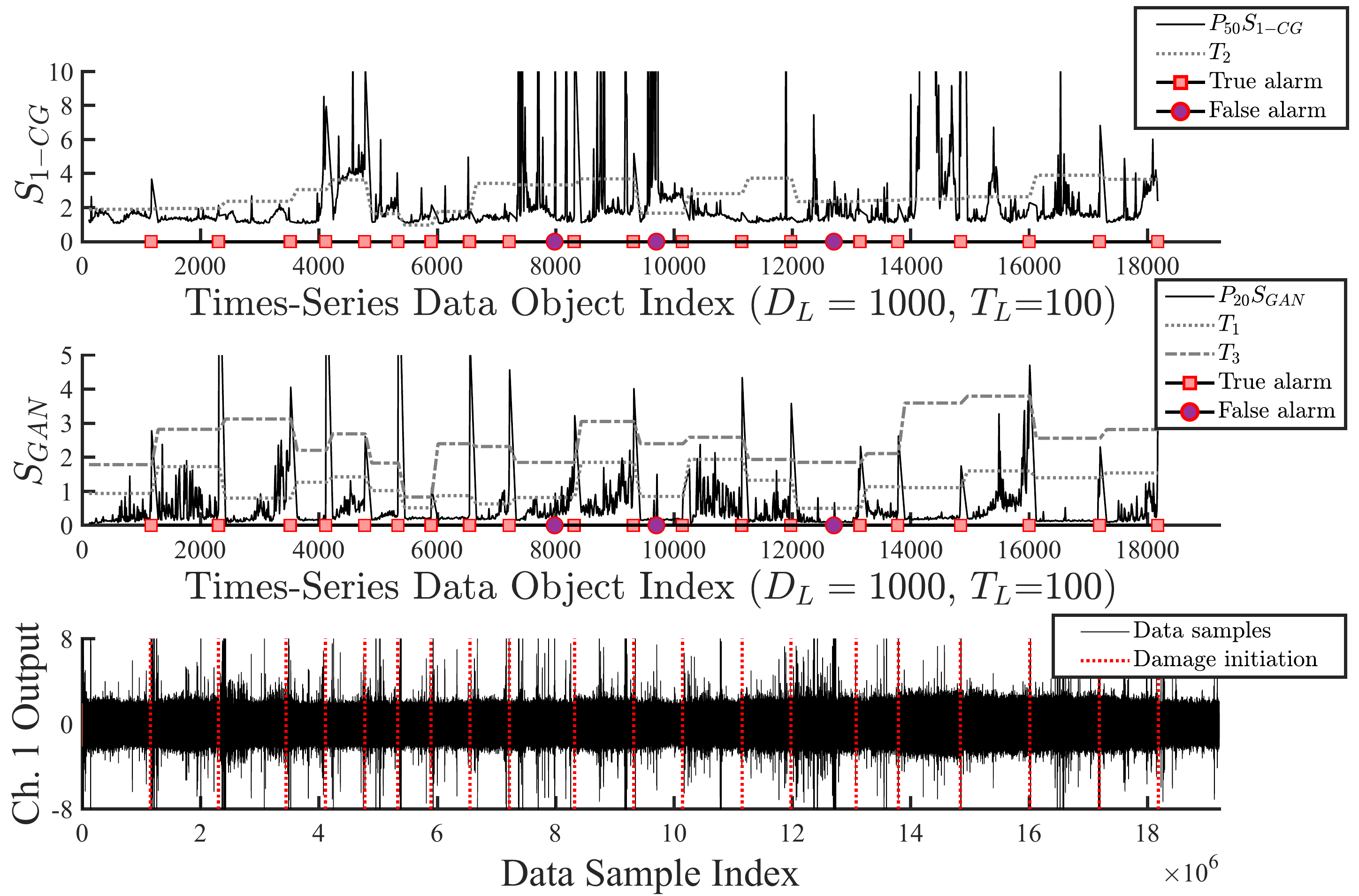}
 \caption {Dynamic baseline novelty detection results with $V_L$ equal to ten, Yellow Frame. (in color)}
 \label{YFWL10}
    \end{figure}

\begin{figure}[h!]
    \centering
     \includegraphics[width=0.9\textwidth]{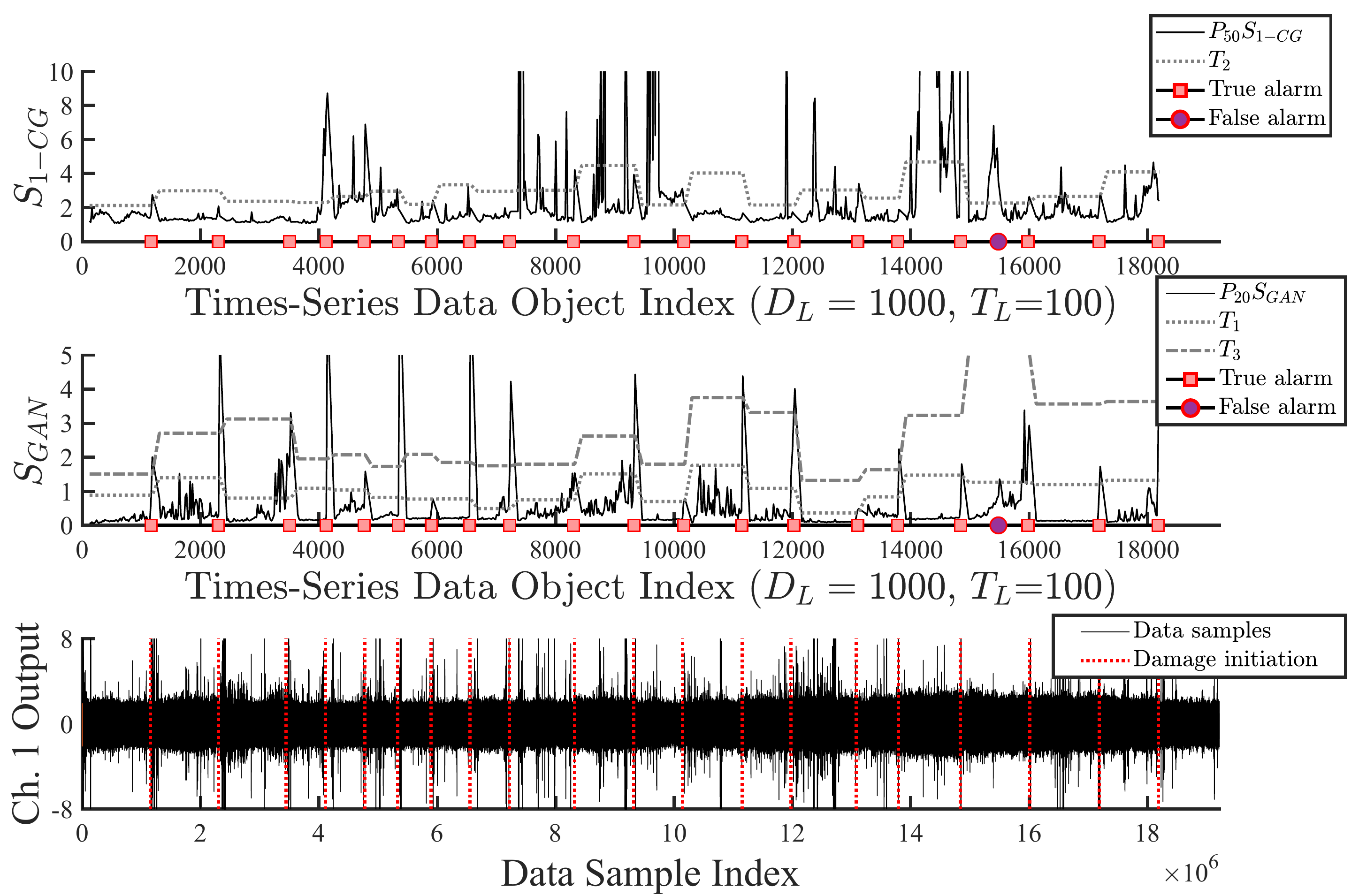}
 \caption {Dynamic baseline novelty detection results with $V_L$ equal to twenty, Yellow Frame. (in color)}
 \label{YFWL20}
    \end{figure}

\begin{figure}[h!]
    \centering
     \includegraphics[width=0.9\textwidth]{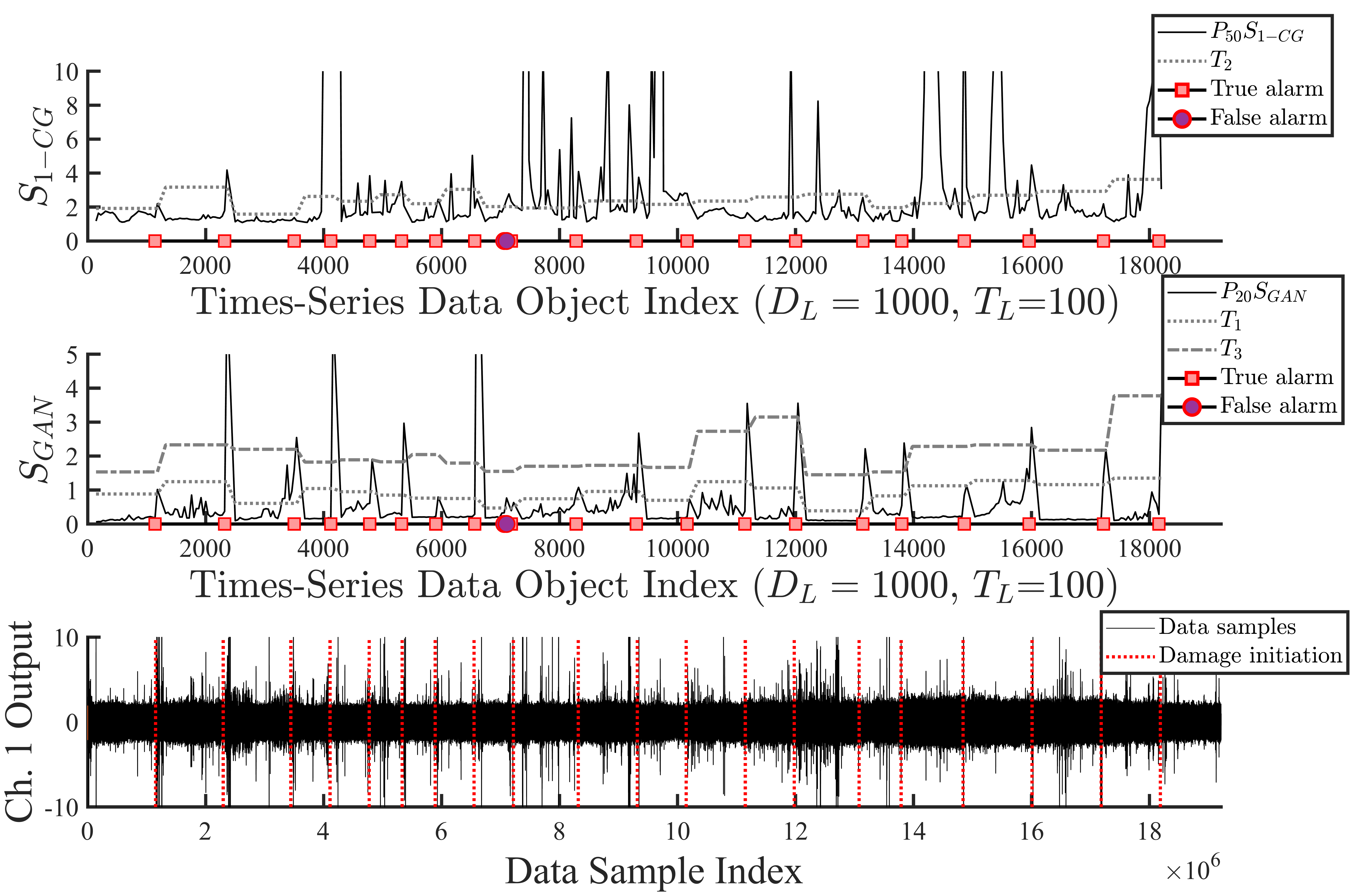}
 \caption {Dynamic baseline novelty detection results with $V_L$ equal to forty, Yellow Frame. (in color)}
 \label{YFWL40}
    \end{figure}

For the three different $V_L$ values of 10, 20, and 40, the results of the unsupervised real-time SHM framework are shown in Fig. \ref{YFWL10}, Fig. \ref{YFWL20}, and Fig. \ref{YFWL40}, respectively. For $V_L=10$, all damages are recognized without delay. There are three false alarms in 1818 vectors of time-series data objects. Hence, the false alarm ratio is equal to 0.165\%. For $V_L=20$, all damages are recognized without delay with one false alarm in 909 vectors of time-series data objects, equivalent to a 0.11\% false alarm ratio. For $V_L=40$, all damages are recognized without delay. There is one false alarm in 404 vectors of time-series data objects, equivalent to a 0.24\% false alarm ratio.
 
The $S_{1-CG}$ scores are noisy (Figs. \ref{YFWL10}, \ref{YFWL20}, and \ref{YFWL40}), as there are plenty of situations in which its detection threshold is exceeded. On the contrary, $S_{GAN}$ is very sensitive to the presence of a novelty, mostly oscillates in a specific range and increasing sharply in the presence of a novelty. Such a trend indicates the potential of high-dimensional features for a robust novelty detection procedure. However, $S_{1-CG}$ also plays an essential role in novelty detection as $T_{1}$ might be too high due to non-deleted outliers in Load histograms or the amount of alteration in the F\_I's features of successive classes of data (e.g., close damage scenarios). In such cases, $S_{1-CG}$ can aid the damage detection.

    
For $V_L=10$ (i.e., 1818 vectors of time-series data objects), the detection results for different $\beta$ values are reported in Table \ref{BEA}. In this table, the false alarm ratios are expected to perceive the $\beta$ variation's effect on the detection results. It is worth mentioning that for higher $\beta$ values, more simulations are required for the MCHS; hence, for this section, all MCHS are performed with 12500 iterations. Based on the main assumption in the tuning phase, $\beta$ should be selected as high as possible, as the analogous system must not fail (i.e., low probability of failure) for the GAN-generated data objects.

Based on Table. \ref{BEA}, low $\beta$ values result in a higher false alarm ratio while higher $\beta$ reduces that ratio. The detection framework's sensitivity to novelties is the most important parameter in the $\beta$ selection. Increasing $\beta$ reduces the false alarms but at the same time increases the detection thresholds, which in turn makes the novelty detection harder. The proposed high-dimensional features, and specifically the GAN's Discriminator, are sensitive to novel data. As an example, in Fig. \ref{YFWL20}, the FDC2's normal data has $S_{GAN}$ of around one, while at the point of novelty detection, it has a $S_{GAN}$ of about six. $S_{GAN}$ is a logarithmic score, which means that the Discriminator's score (i.e., probability) for a novel data object is around million times smaller than a normal data object's score. This characteristic enables selecting a high $\beta$, which can ensure less false alarm ratio while maintaining the novelty detection power. There are instances of delay in some data classes with high $\beta$ values (Table \ref{BEA}). Due to the large thresholds caused by high $\beta$ values, the detection may not occur at the point of the damage initiation; however, with some additional iterations and the fluctuation of scores, the damage is finally detected.

\begin{table}[h!]
  \caption{Detection summaries for Yellow Frame with $WL=10$, and different $\beta$ values.}
  \label{BEA}
  \centering
  \begin{tabular*}{\linewidth}{@{\extracolsep{\fill}} l ccc}
    \toprule
$\beta$ &False alarms& False alarm ratio (\%) & Novelty detection results \\
    \midrule
1&	90&4.95&	All detected\\	
2&	20&	1.10&	All detected\\	
3&	3&0.16&	All detected\\	
3.5	&3	&0.16&	All detected\\
4&	3&	0.16&All detected, Delay	in FDC16\\
5&	2&	0.11&	All detected, Delay	in FDC16 and FDC7\\
\bottomrule
\end{tabular*}
\end{table}
 
\begin{figure}[h!]
\centering
\includegraphics[width=0.9\textwidth]{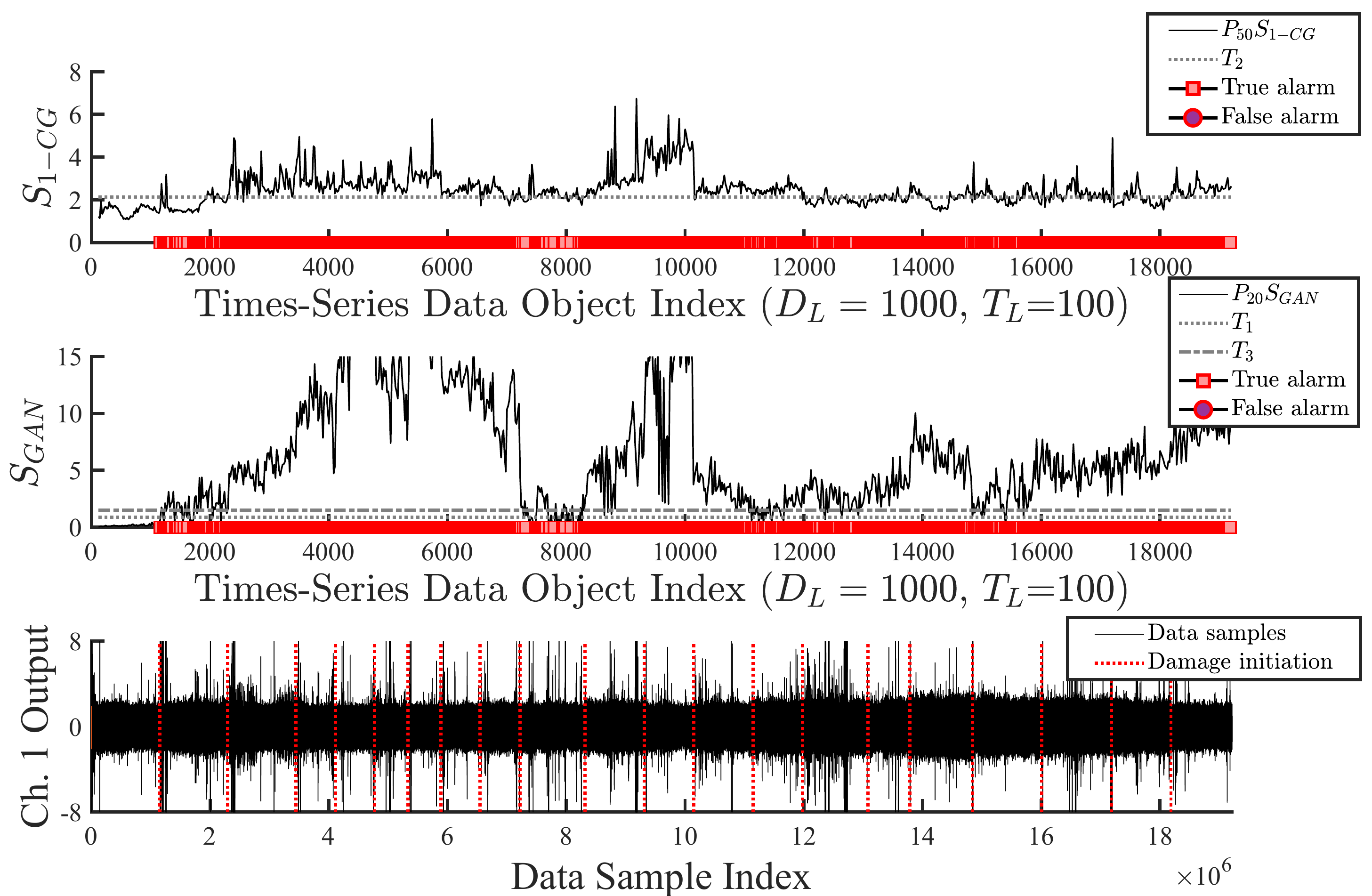}
 \caption {Static baseline novelty detection results with $V_L$ equal to twenty, Yellow Frame. (in color)}
\label{YFWLA}
\end{figure}

The novelty detection is also applied in the format of static baseline,  with $V_L=20$. The GAN and 1-CG models are trained only for the normal state, and the framework is used to detect all damage classes as a single novelty class. The result is shown in Fig. \ref{YFWLA}. Based on this figure, since only one true alarm at the beginning of each damage scenario is enough for that class to be considered detected without delay, all damages are identified with no false alarms. Although both static and dynamic baseline methods performed well, the dynamic baseline is of the main interest in this study as having a static number of classes in an unsupervised SHM is a drawback. However, in cases where the binary classification is the main concern, the static baseline approach can be used considering the Yellow Frame datasets' reliable detection results with 21 data classes.

 \section{Z24 Bridge results and discussion}
 
The Z24 Bridge dataset is utilized to evaluate the proposed method's generalizability to bridges. For this dataset, and to show the reliability-based unsupervised real-time SHM approach's capability of tuning the detection thresholds to avoid detection sensitivity to the user-defined parameters, three values of five, ten, and twenty-five are chosen for $V_L$. $T_L$ is also equal to seventy-five. In what follows, results from each portion of the proposed unsupervised real-time SHM method (Fig. \ref{M1}) are presented and discussed. 

\subsection{GAN and 1-CG training, Z24 Bridge}

\begin{figure}[h!]
    \centering
    \begin{subfigure}[b]{0.48\linewidth} 
        \centering
     \includegraphics[width=\textwidth]{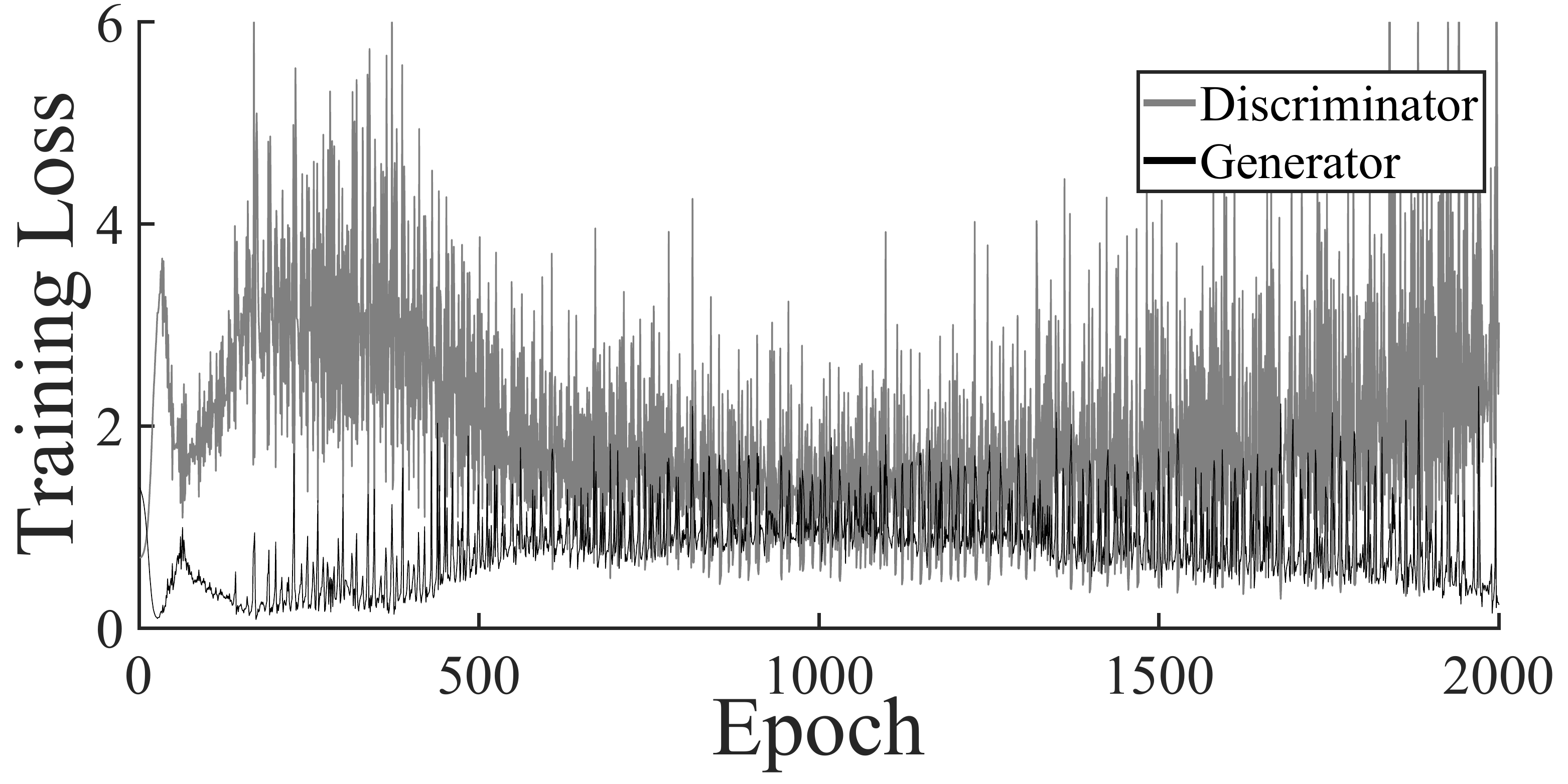}
        \caption {}
        \label{fig:GZ}
    \end{subfigure}
    \begin{subfigure}[b]{0.48\linewidth}
        \centering
       \includegraphics[width=\textwidth]{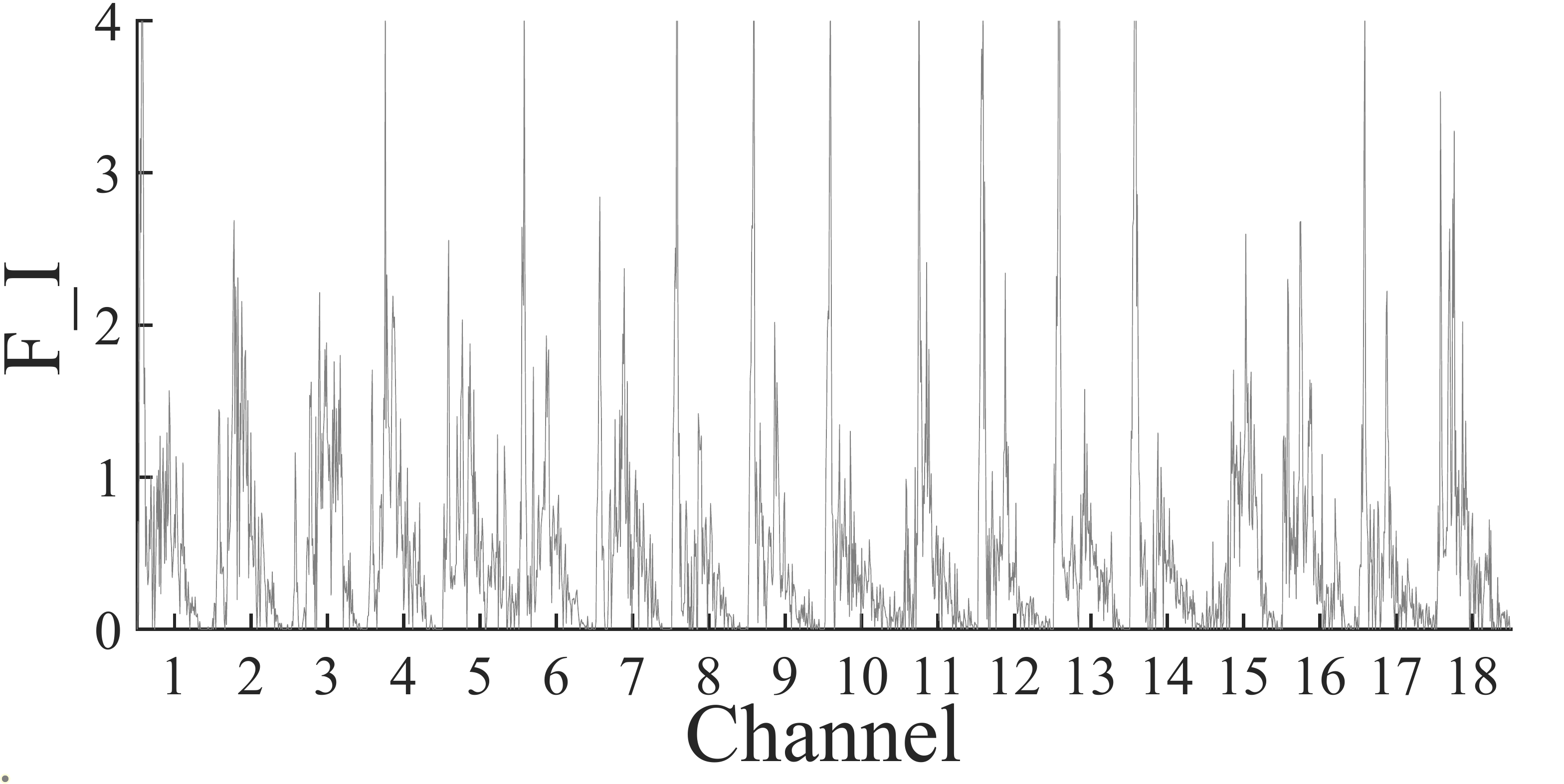}
        \caption{}
        \label{fig:GZ2}
    \end{subfigure}
    \caption{Unsupervised real-time SHM framework results for BDC1 (a) the GAN' training loss (b) a generated F\_I feature.}
    %\label{fig:approach1}
\end{figure}

The Z24 data objects' F\_I features have a dimension of $1800\times1$. As mentioned, this dimension is less than the Yellow Frame F\_I features dimension (i.e., $7500\times1$); hence, the GAN's training is conducted with 2000 epochs. The F\_II feature for this dataset has the dimension of $54\times1$ and is extracted to train the 1-CG models. The GAN's training loss and a sample generated F\_I feature for the BDC4 are shown in Fig. \ref{fig:GZ} and Fig. \ref{fig:GZ2}, respectively. The 1-CG models are also trained with the training set to be used for tuning the detection thresholds.

\subsection{Novelty-detection system's tuning for $V_L$}

\begin{figure}[h!]
    \centering
    \begin{subfigure}[b]{0.32\linewidth} 
        \centering
     \includegraphics[width=\textwidth]{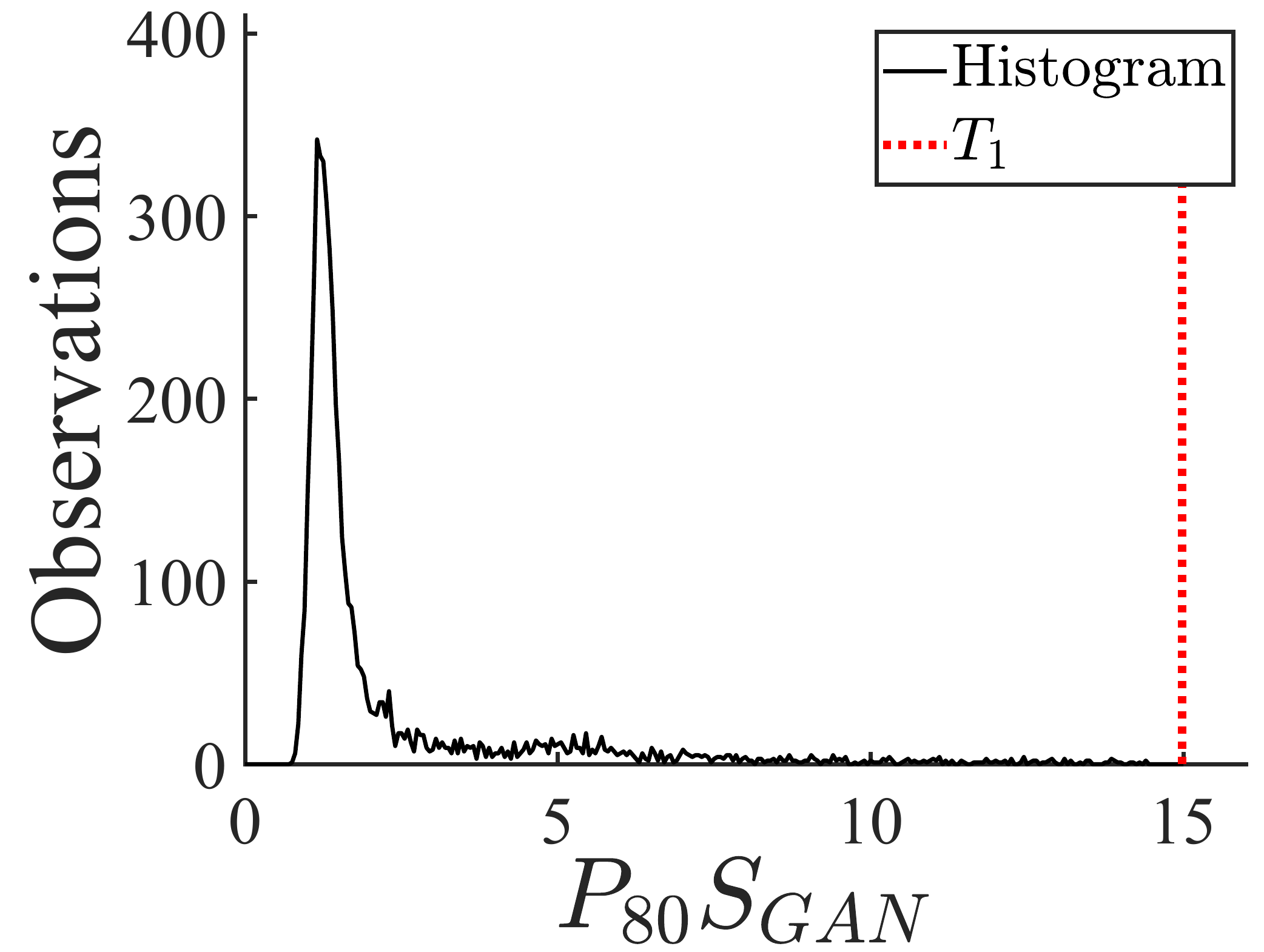}
        \caption {}
        \label{fig:GZO}
    \end{subfigure}
    \begin{subfigure}[b]{0.32\linewidth}
        \centering
       \includegraphics[width=\textwidth]{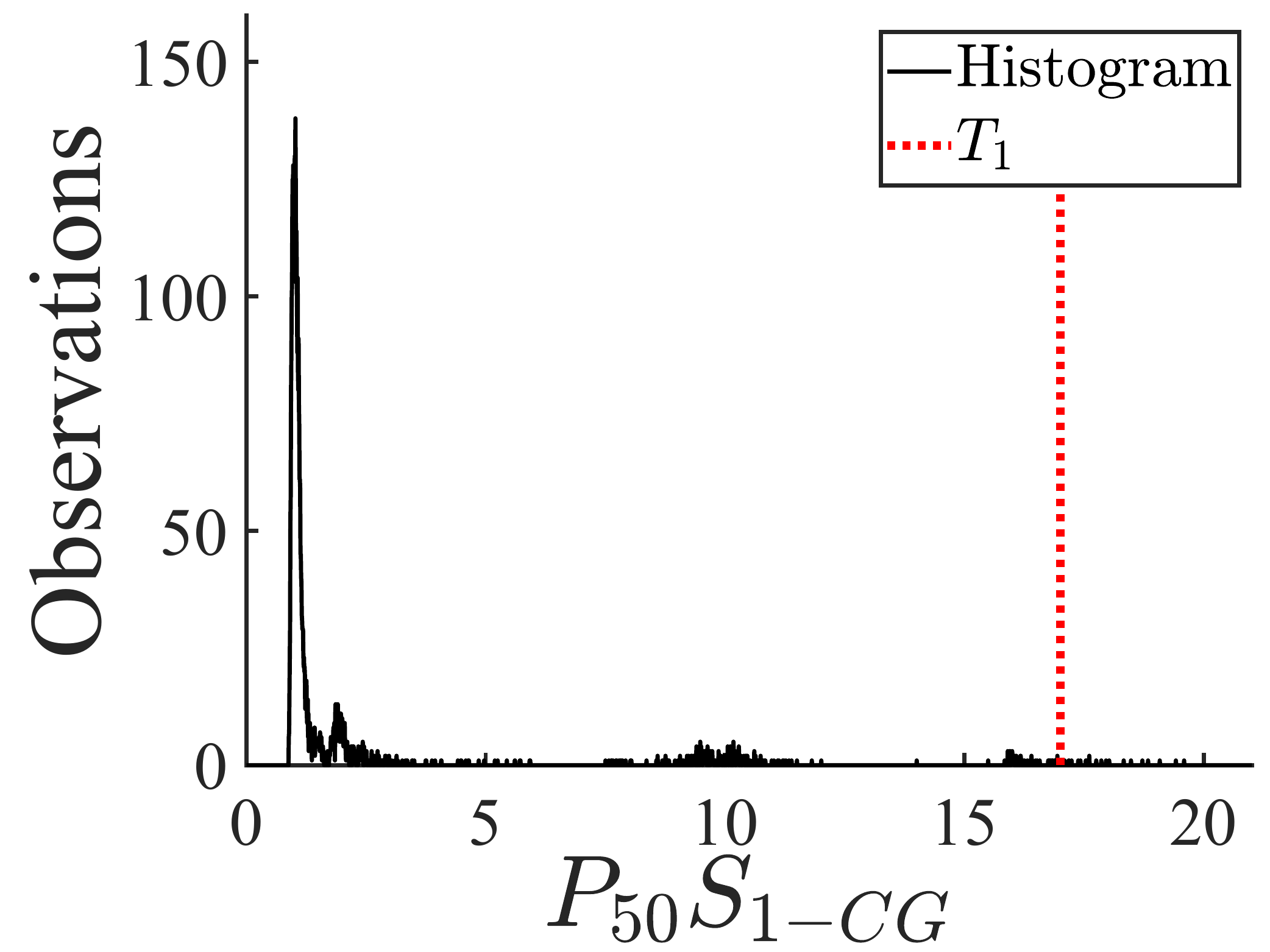}
        \caption{}
        \label{fig:GZ2O}
    \end{subfigure}
        \begin{subfigure}[b]{0.32\linewidth}
        \centering
       \includegraphics[width=\textwidth]{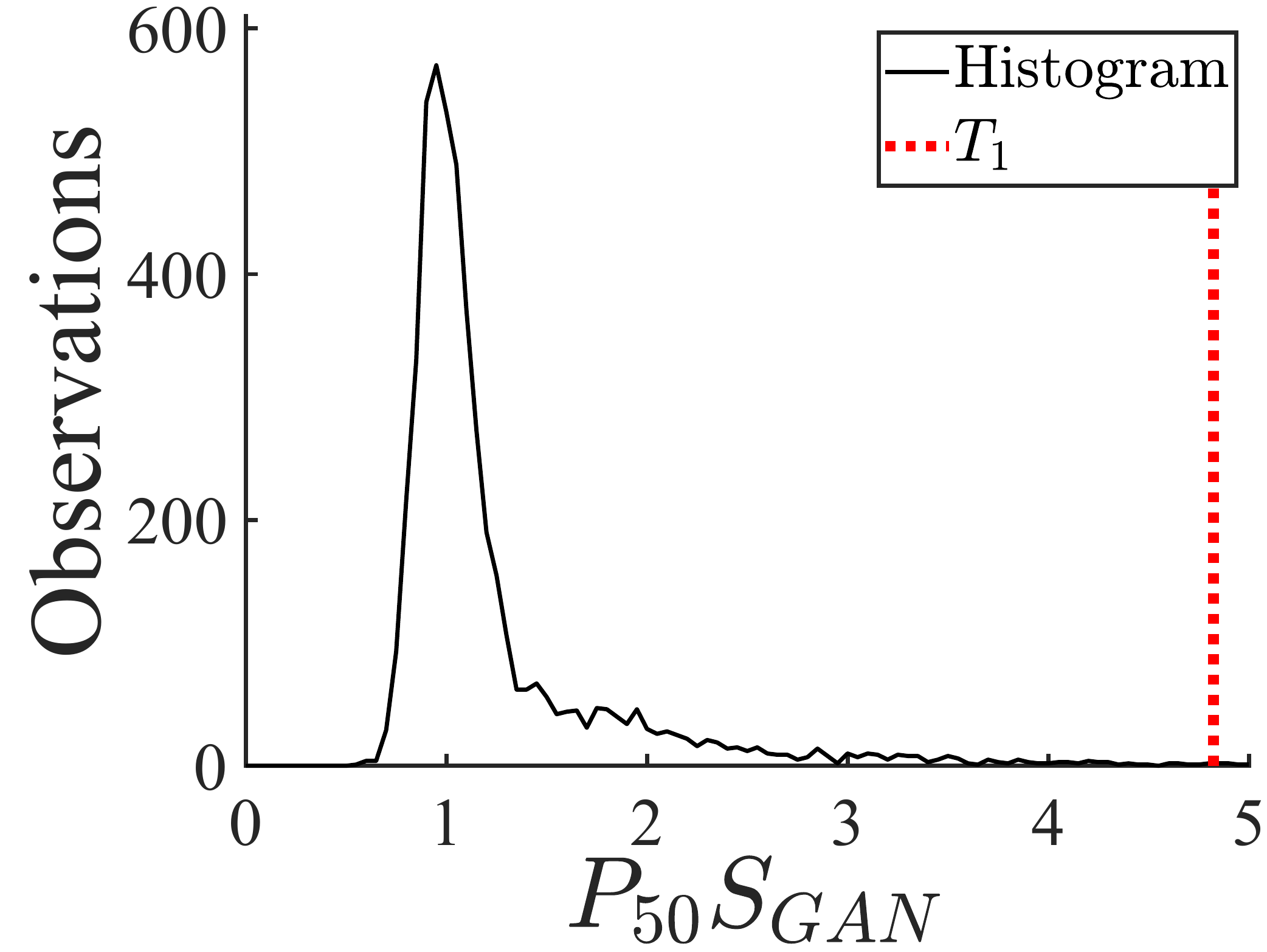}
        \caption{}
        \label{fig:GZ3O}
    \end{subfigure}
    \caption{Load MCHS results of novelty-detection system's Elements (a) I (b) II (c) III.}
    \label{BOZ}
    %\label{fig:approach1}
\end{figure}

\begin{figure}[h!]
    \centering
    \begin{subfigure}[b]{0.32\linewidth} 
        \centering
     \includegraphics[width=\textwidth]{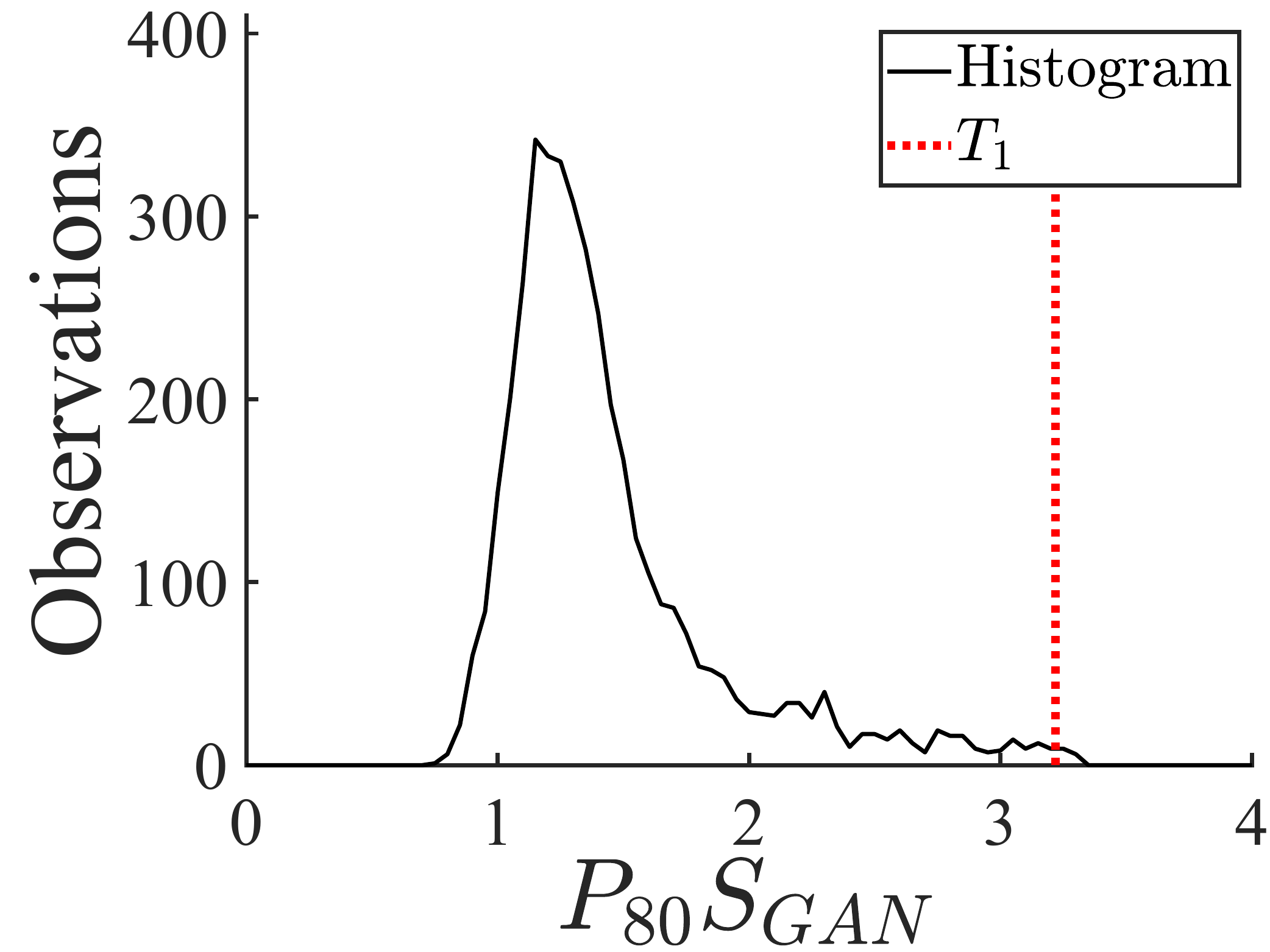}
        \caption {}
        \label{fig:GZW}
    \end{subfigure}
    \begin{subfigure}[b]{0.32\linewidth}
        \centering
       \includegraphics[width=\textwidth]{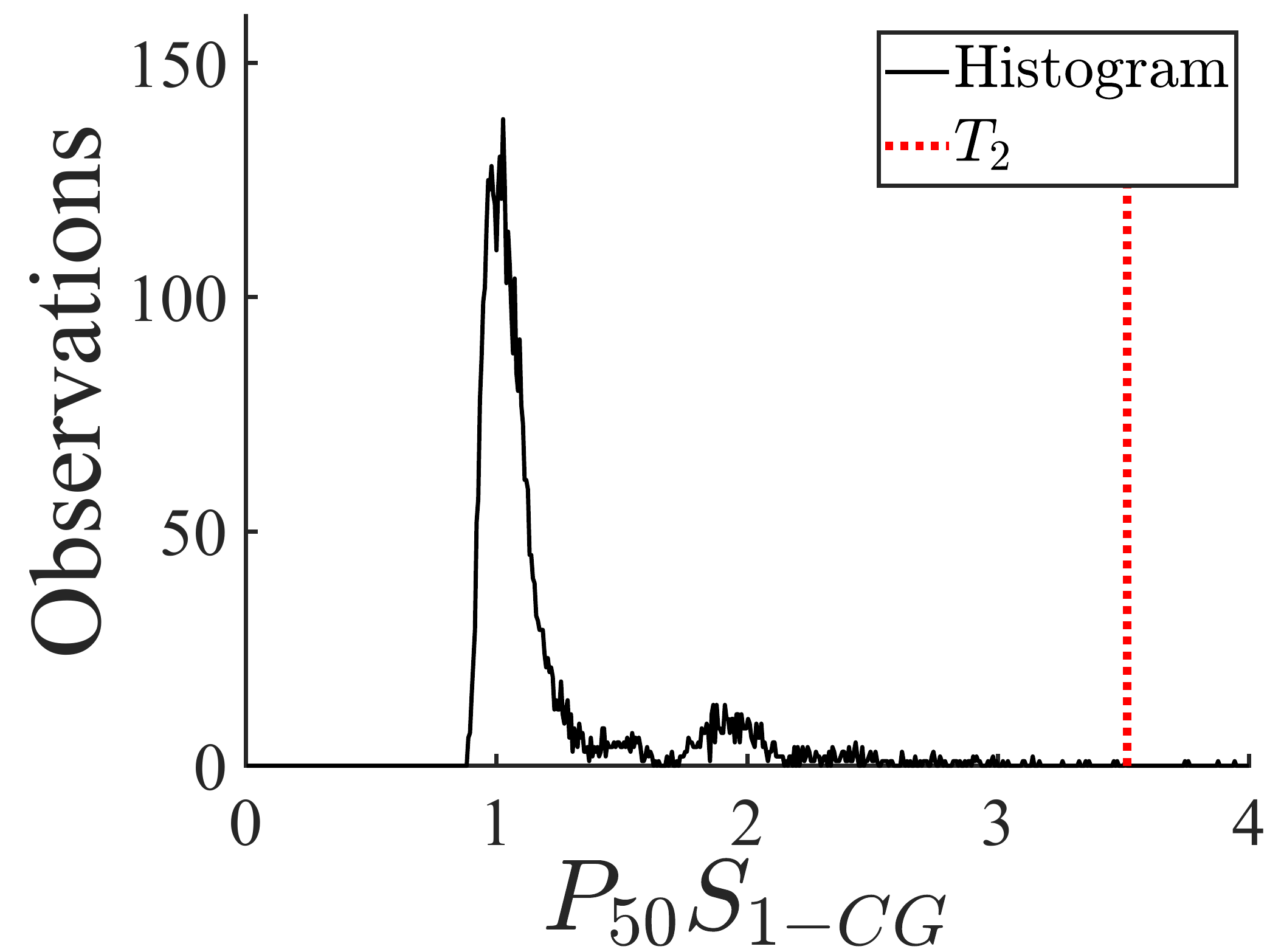}
        \caption{}
        \label{fig:GZ2W}
    \end{subfigure}
        \begin{subfigure}[b]{0.32\linewidth}
        \centering
       \includegraphics[width=\textwidth]{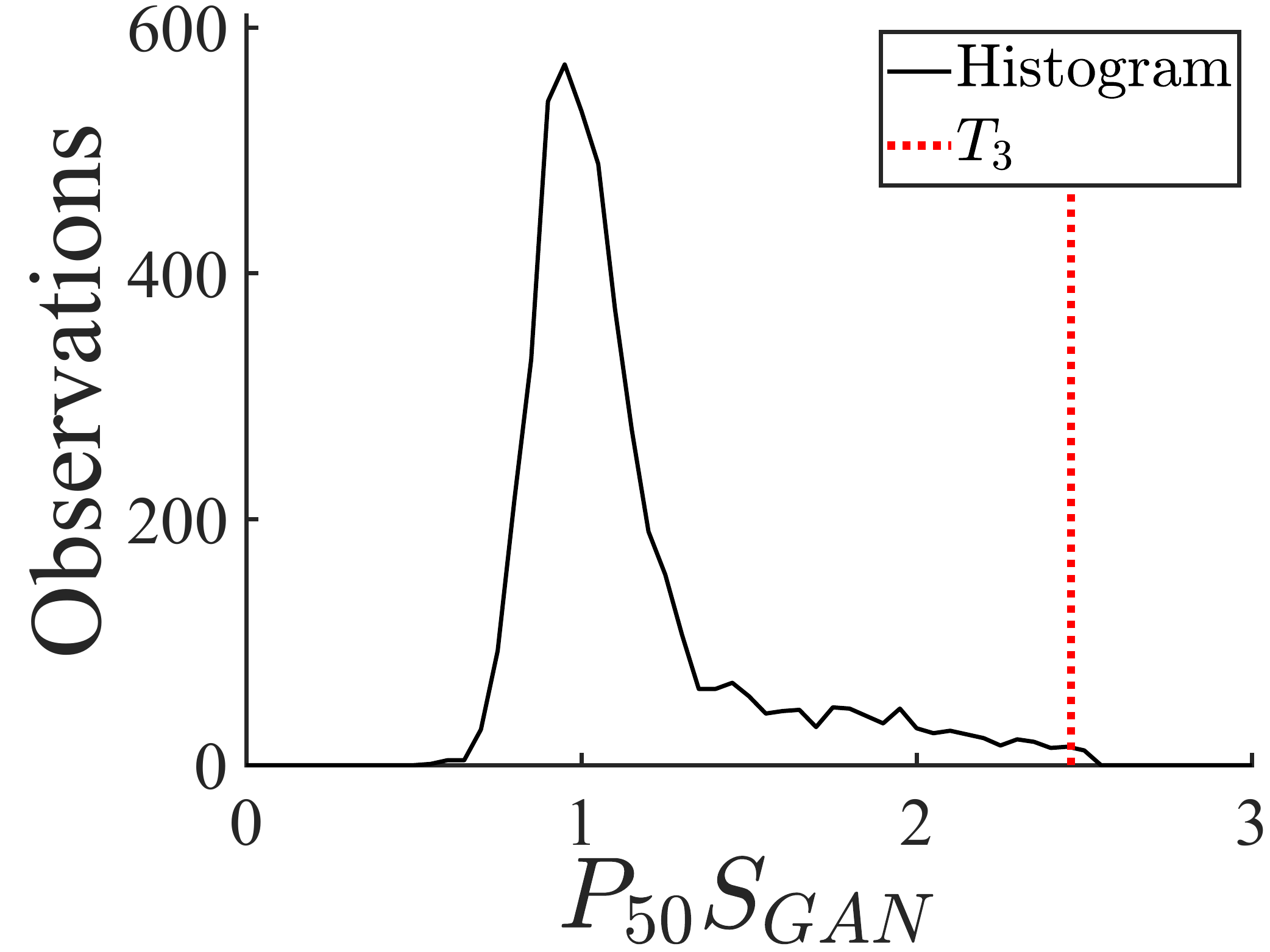}
        \caption{}
        \label{fig:GZ3W}
    \end{subfigure}
    \caption{MCHS results (without outliers) of novelty-detection system's Elements (a) I (b) II (c) III.}
    \label{AOZ}
    %\label{fig:approach1}
\end{figure}

With the trained Generator, Discriminator, and 1-CG model, MCHS of Load for different novelty detection systems' elements is performed to define the detection thresholds. All three Load histograms before and after the application of 2-CGMM (i.e., outlier-detector) for BDC1 is shown in Fig. \ref{BOZ}, and Fig. \ref{AOZ}, respectively. The Z24 Elements' Load histograms tend to be much noisier than the Yellow Frame ones. Without the application of 2-CGMM (Alg. \ref{Alg1}), the thresholds would be so large, and the chance of missing novelties is higher. Unlike the Yellow Frame case, all three histograms have changed by the application of 2-CGMM.

\subsection{Real-time novelty detection results, Z24 Bridge}

In what follows, results from each segment of the unsupervised real-time SHM (Fig. \ref{M1}) are presented and discussed. Although Z24 has fourteen main damage classes, some damages are intensified counterparts of a preceding damage scenario. Following the same procedure as in the Yellow Frame case, a novelty is labeled as a false alarm if initiated before starting a new class of damage minus $V_L$. Both dynamic and static baseline approaches (Fig. \ref{M1}) are applied, and results are discussed.

\begin{figure}[h!]
    \centering
     \includegraphics[width=0.9\textwidth]{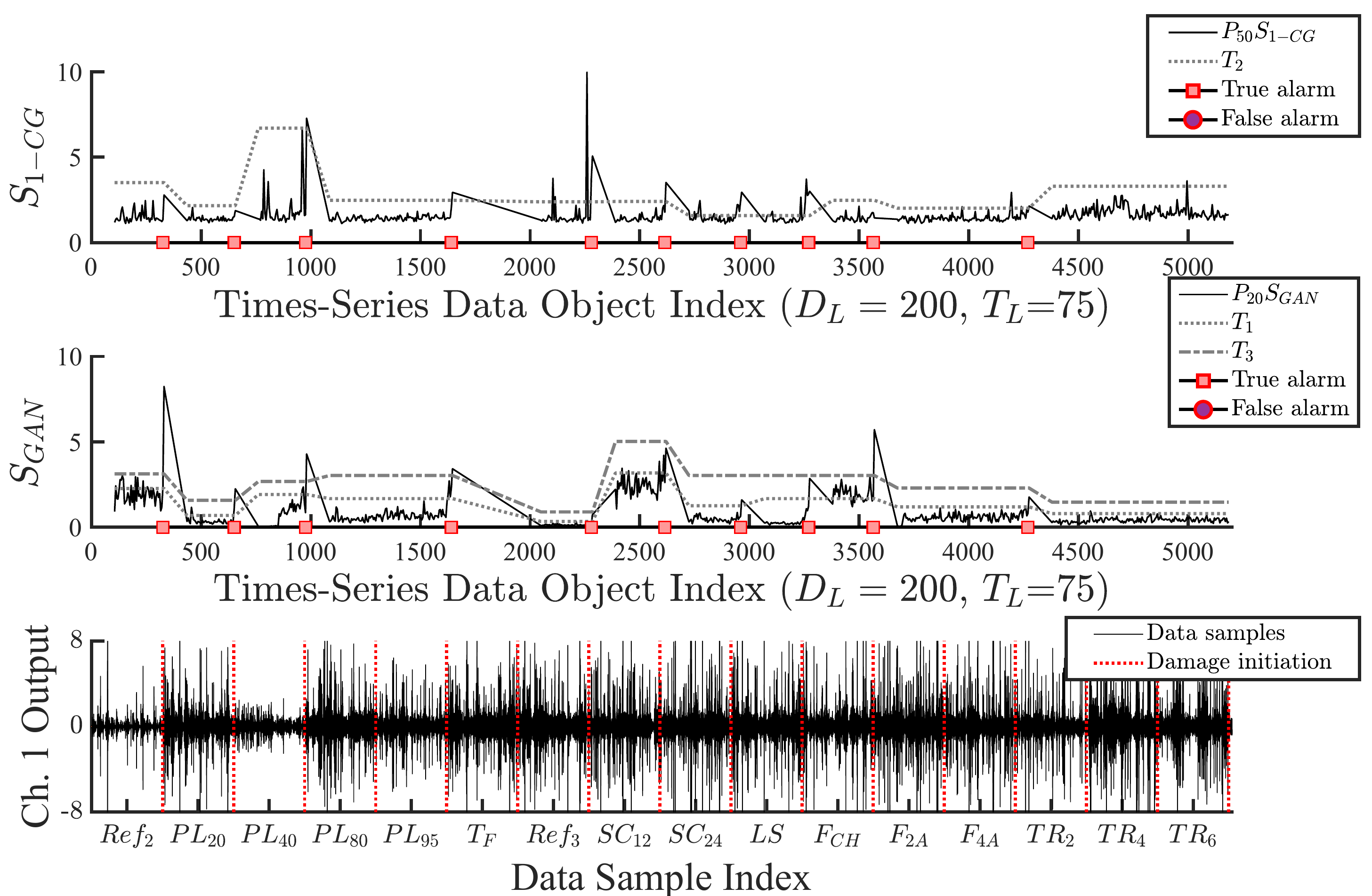}
 \caption {Dynamic baseline novelty detection results with $V_L$ equal to five, Z24. (in color)}
 \label{Z245}
    \end{figure}

\begin{figure}[h!]
    \centering
     \includegraphics[width=0.9\textwidth]{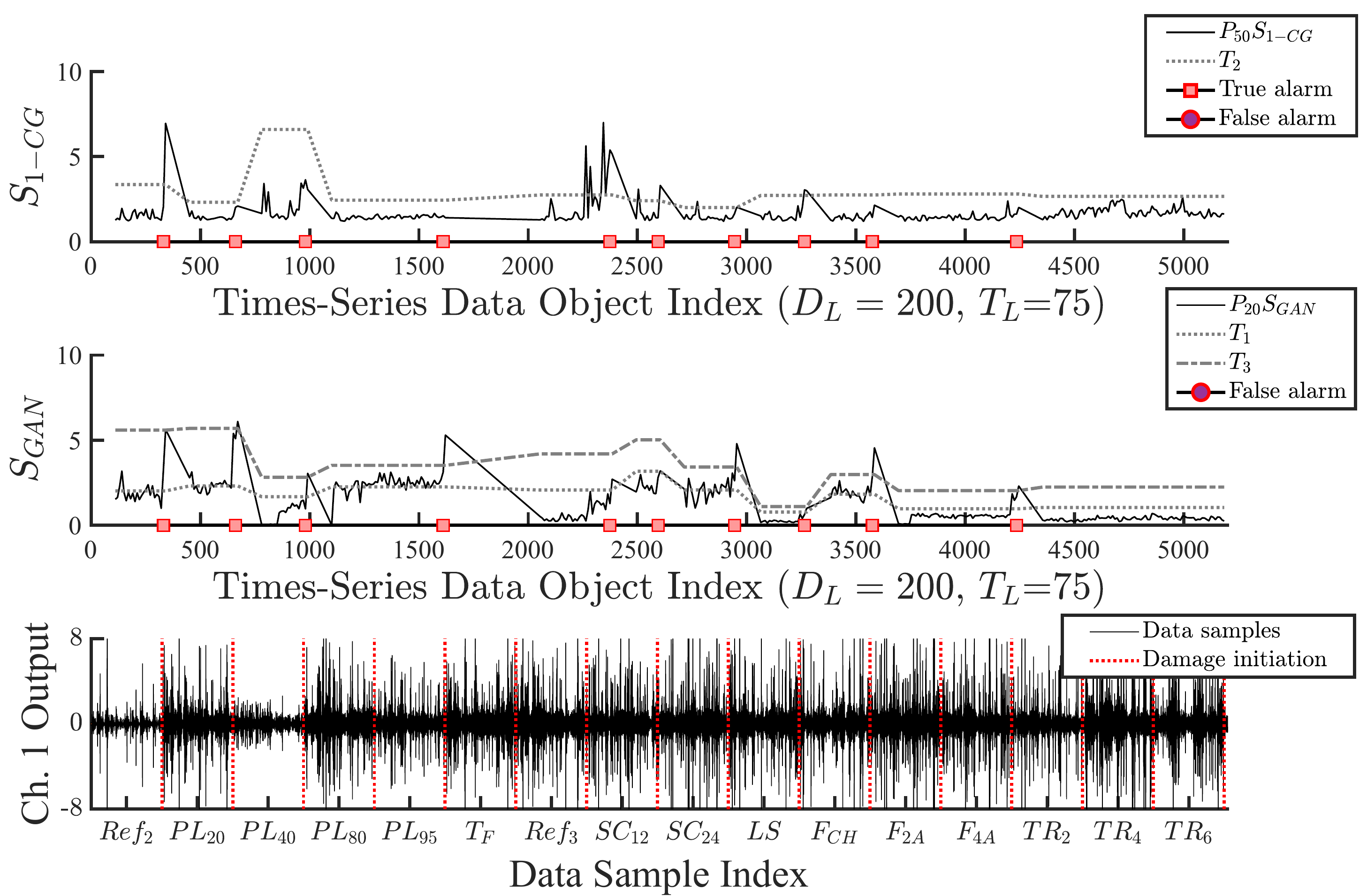}
 \caption {Dynamic baseline novelty detection results with $V_L$ equal to ten, Z24. (in color)}
 \label{Z2410}
    \end{figure}

\begin{figure}[h!]
    \centering
     \includegraphics[width=0.9\textwidth]{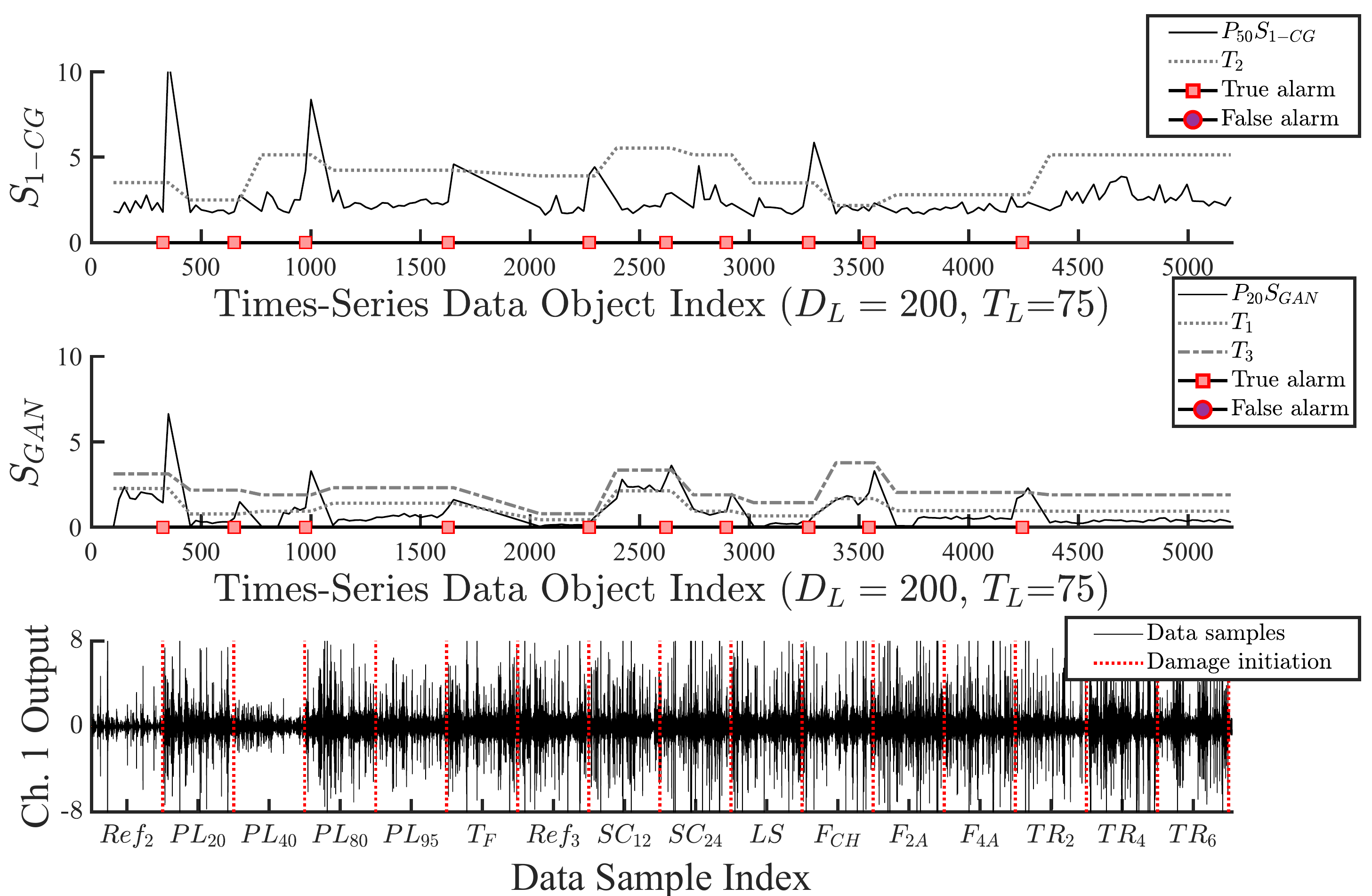}
 \caption {Dynamic baseline novelty detection results with $V_L$ equal to twenty-five, Z24. (in color)}
 \label{Z2425}
    \end{figure}

 The dynamic baseline detection results are shown in Fig. \ref{Z245}, Fig. \ref{Z2410}, and Fig. \ref{Z2425} for $V_L$ of 5, 10, and 25, respectively. With different $V_L$ values, all seven distinct damage scenarios (i.e., lowering of the pier, tilt of foundation, spalling, landslide, failure of hinges, and tendon ruptures) are identified with no false alarms. These seven main classes of damage are on different scales, global or local; hence, their detection shows the proposed framework's versatility in damage detection. The results indicate that for some successive damage scenarios with different intensities, detection does not take place. The detection between 95mm lowering of the pier from 80mm of the pier lowering or failure of four anchor heads from two anchor heads' failure has not happened. Furthermore, the damage class of six ruptured tendons is not discerned from the class of four ruptured tendons. However, some successive different damage-intensity categories are captured, such as the 40mm and 80mm of pier lowering from 20mm and 40mm of pier lowering or different concrete spalling areas. This phenomenon is reasonable, as different tendon ruptures may change a specific part of the FFT of sensor signals, making it hard for GAN to capture the difference.

\begin{figure}[h!]
    \centering
     \includegraphics[width=0.9\textwidth]{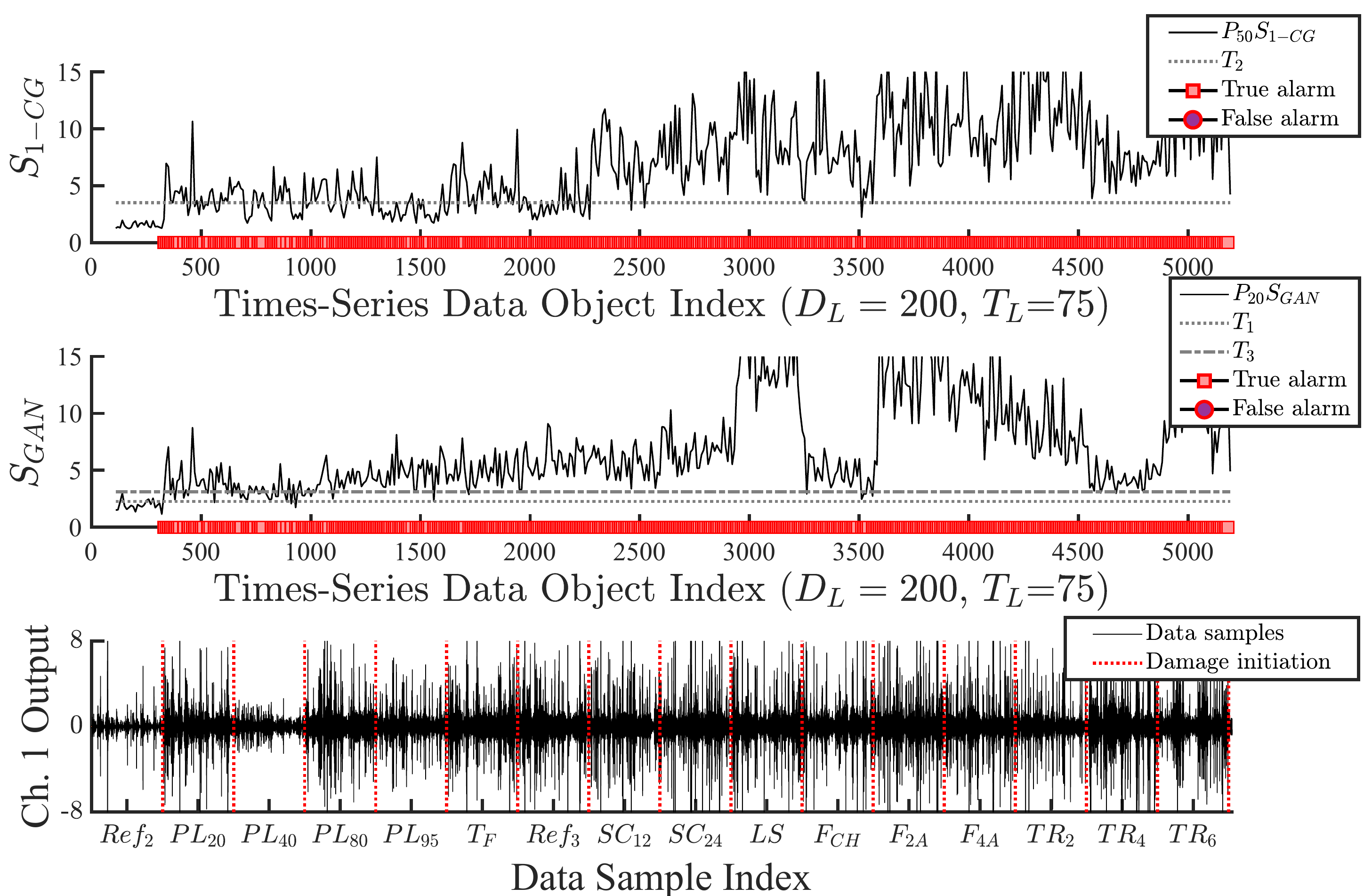}
  \caption {Static baseline novelty detection results with $V_L$ equal to ten, Z24. (in color)}
 \label{YFWLAAA}
    \end{figure}

The static baseline approach is also applied to the dataset. The result is shown for $V_L=10$ in Fig. \ref{YFWLAAA}. The results show that all different types of damages, with varying intensities, are detected by the framework, which can ensure (in this study) that the method can detect all damage classes, even with the same origin and different intensities from the normal state of the structure.

\section{Future work}

The proposed method is a novel framework with many possible areas for improvement. Various network architectures for GAN's Generator and Discriminator can be investigated, with convolutional or RNN layers. The novelty detection system also has room for further research, as with the ensemble of GAN and 1-CG, different and more suitable systems can be employed. Performing MCHS or other reliability analysis methods on the whole novelty detection system without simplifying assumptions is another topic. Furthermore, all GAN pieces of training are done without any supervision and with a fixed number of epochs. GAN has no specific concluding training epoch; hence, different training epochs (i.e., training loss diagram patterns) can be checked for possible shortcomings or improvements in the results.

\section{Conclusion}

Investigating unsupervised, data-driven methods as the most suitable approach to perform real-time structural health monitoring has received so much attention in the last decade. There are many improvements in the field, and this study tries to handle four of the main obstacles in the current approaches. The obstacles are loss of information from dimensionality reduction, case-dependency of feature extraction steps, lack of dynamic clustering, and the effect of user-defined parameters on the novelty detection results. Two low- and high-dimensional FFT-based features apply to any instrumented structure (i.e., having sensors installed) are defined. A multi-ensemble of joint Gaussian distribution (1-CG) models for the low-dimensional feature and GAN for the high-dimensional feature is established. A novelty detection system of limit-state functions is defined with the detection thresholds as Resistance and scores from trained GAN and 1-CG models as Loads. Taking advantage of trained GAN's Generator, which generates data objects from a standard Gaussian distribution, the novelty detection system's thresholds can be tuned to user-defined parameters. The Monte Carlo histogram sampling approach is utilized for the tuning. For depicting the proposed method's generalization capacity, two datasets of a frame and a bridge are used. The Yellow Frame dataset has twenty-one classes, while Z24 Bridge has fifteen data classes. Applying an unsupervised real-time method on such high-class datasets is scarce in the literature. Based on the results, the following conclusions are made.
 
The framework's application in a static baseline setting resulted in no false alarms on both Yellow Frame and Z24 datasets, and all twenty damage classes in Yellow Frame and fifteen damage classes in Z24 were identified. A dynamic baseline approach identifies different novelties from the normal state and other damage classes (i.e., multi-class). The dynamic baseline approach to Yellow Frame resulted in detecting all twenty changes in the structural condition. The false alarms were three, one, and two for $V_L$ of 10, 20, and 40, respectively. The resulting false-alarm ratios show that the reliability-based thresholds' tuning made the detection results insensitive to user-defined detection parameters (i.e., $V_L$). The dynamic baseline approach's for Z24 resulted in the detection of all seven distinct damage classes. In cases of different intensities of damage, such as 95mm and 80mm lowering of the pier, or between different numbers of ruptured tendons, the method could not catch the difference. The reason might be that different intensity from the same damage scenario affects the FFT of signals in the same way. Hence, the detection models cannot tell those classes' difference. However, detecting different damage intensities is not as important as alarming new damage or damage following an intact condition. All the results originated from a simple feature-extraction step, i.e., the half-spectrum FFT. The simple extraction method can be applied to new structures without any training phase. The simple application characteristic is a vital step towards having a network of monitored facilities on big scales as a part of futuristic smart cities.
 
 \section*{Acknowledgement}
The authors would like to thank Carlos Ventura, Alex Mendler, and Saeid Allahdadian for providing the Yellow Frame dataset and associated images. The structural mechanics section of KU Leuven is gratefully acknowledged for providing access to the Z24 Bridge dataset.

 \section*{Declaration of conflicting interests}
The author(s) declared no potential conflicts of interest concerning the research, authorship, and/or publication of this article.

\bibliography{Bib.bib,library.bib}

\end{document}